\documentclass{article}

\usepackage{arxiv}

\usepackage[utf8]{inputenc} 
\usepackage[T1]{fontenc}    
\usepackage{hyperref}       
\usepackage{url}            
\usepackage{booktabs}       
\usepackage{amsfonts}       
\usepackage{nicefrac}       
\usepackage{microtype}      
\usepackage{lipsum}
\usepackage{graphicx}

\usepackage{amsthm}

\usepackage{amsmath}
\usepackage{amssymb}
\usepackage{bbm}
\DeclareMathOperator*{\E}{\mathbb{E}}
\usepackage{tablefootnote}
\theoremstyle{definition}
\newtheorem{definition}{Definition}
\newtheorem{example}{Example}

\title{Hide-and-Seek: A Template for Explainable AI}

\author{
  Thanos Tagaris \thanks{Code for this paper: \texttt{https://github.com/djib2011/hide-and-seek}}\\
  Artificial Intelligence and \\Learning Systems Laboratory,\\
  National Technical University of Athens\\
  \texttt{thanos@islab.ntua.gr} \\
   \And
 Andreas Stafylopatis \\
  Artificial Intelligence and \\Learning Systems Laboratory,\\
  National Technical University of Athens\\
  \texttt{andreas@cs.ntua.gr} \\
}

\begin{document}
\maketitle

\begin{abstract}
Lack of transparency has been the Achilles heal of Neural Networks and their wider adoption in industry. Despite significant interest this shortcoming has not been adequately addressed. This study proposes a novel framework called Hide-and-Seek (HnS) for training Interpretable Neural Networks and establishes a theoretical foundation for exploring and comparing similar ideas. Extensive experimentation indicates that a high degree of interpretability can be imputed into Neural Networks, without sacrificing their predictive power.
\end{abstract}


\section{Introduction}

Neural Networks have seen a sharp increase in popularity over the past years, mainly due to the emergence of Deep Learning techniques \cite{lecun2015deep}.
They have benefited from intense research interest, which helped them make leaps in several tasks including computer vision \cite{DBLP:journals/corr/abs-1906-06423, DBLP:journals/corr/abs-1812-01593} and natural language processing \cite{lample2018phrase, DBLP:journals/corr/abs-1906-08237}. 
These networks, which have surpassed several milestones would have been considered unreachable a few years ago and they have shown no indication of slowing down.


Regardless of the explosion in popularity and their remarkable performance, Neural Networks (NN) have experienced a relatively slow adoption rate in several fields of science \cite{kang2015machine, mamoshina2016applications}. 
For example, while Neural Networks have achieved significant results in medical imaging \cite{alom2019state, biswas2019state, kollias2018deep, plis2014deep} and drug discovery \cite{vamathevan2019applications} and in some cases have even surpassed human performance \cite{de2018clinically}, their practical applications remain limited.
The most prominent domains which suffer from this are healthcare and autonomous driving.

This slow adoption can be attributed to a combination of practical \cite{papernot2017practical, su2019one}, social \cite{angwin2016, DBLP:journals/corr/BolukbasiCZSK16a, Baeza-Yates:2018:BW:3229066.3209581} and ethical reasons \cite{Allen2005, lin2017robot, 7924235, 1667947, anderson2011machine}. 
Some of these fall out of the scope of Machine Learning (ML), as they mostly involve legal accountability (e.g. \textit{who is to blame if a self-driving car crashes?}), adverse side-effects (e.g. jobs becoming redundant) or philosophical questions on machine morality.

Still, a plethora of reasons stem from the very nature of Machine Learning. Most of these have a common culprit: the lack of interpretability of ML algorithms \cite{lipton2016mythos, zhang2018visual}. Neural Networks, in particular, are considered as \textit{black boxes}, i.e. one can only observe the inputs and the outputs of the network, but none of the decision-making. In consequence, interpretability is one of the focal points of research in the field of Neural Networks and ML in general.
An intelligent system that can't provide reasoning for how it derives its predictions can't be fully trusted by its users, which will be reluctant to employ that system. This, in fact, has been the case in multiple fields  \cite{linegang2006human, stubbs2007autonomy, miller2018explanation}. 
The running hypothesis is that with more transparent, interpretable or explainable systems users will be better equipped to understand and therefore trust the intelligent agents \cite{chen2014situation, mercado2016intelligent, hayes2017improving}.

Besides improving the trust between user and machine, interpretability can assist in other aspects, as well. These include helping identify the cause of a model's failure \cite{hoiem2012diagnosing} and utilizing models operating at super-human performance teach humans how to make better decisions \cite{johns2015becoming}. Finally, governments around the world have shown intentions in drafting regulations regarding interpretability in Artificial Intelligence (AI) systems. One key example is Article 15 from the EU's GDPR \cite{gdpr}, which states that if a person's data has been subject to a decision-making AI, he has the right to access ``meaningful information about the logic involved". 

Visual models especially suffer from this, as the state-of-the-art networks are getting increasingly more complex. The most influential architectures in computer vision, Convolutional Neural Networks (CNN), are far from perfect. Either through identifying noise-examples that are classified as objects with a high degree of confidence \cite{szegedy2013intriguing}, or through modifying a single pixel in the input image to shatter a CNN's performance (again with high confidence) \cite{su2019one}, researchers have shown how sensitive CNNs actually are. Examples such as this, feed into the reluctance of the various industries to adopt such models, despite their numerous successes. Thus the importance of providing a reasoning for their predictions is paramount to their widespread adoption.

The present study aims to ``shed some light" into the inner-workings of Neural Networks. More specifically, the goal is to develop a framework that serves as a fully-interpretable image classifier. A more detailed explanation of the goal will be provided in Section \ref{sec:definition}, while Section \ref{sec:hns} will describe the proposed framework in detail.

\section{Problem Definition}
\label{sec:definition}

Naturally, due to the increasing interest, a lot of research has been conducted on improving the ``transparency" of Neural Networks and especially CNNs. In the core of this pursuit lies the \textit{fidelity-interpretability trade-off} \cite{ribeiro2016should, selvaraju2017grad}. A simplified explanation is that the increase of a model's capacity (which is necessary for it to solve complex tasks) leads to the decrease of its interpretability. An example of this would a simple 2-layer Neural Network. The addition of the hidden layer, allows for the modelling of non-linear functions but abolishes the direct mapping from input to output, thus making it harder to tell which input affects the network the most. While these two terms are theoretical constructs, an effort will be made to ``ground" them by re-defining them so that they can be quantified. 

The term \textit{Interpretability} on its own is vague and is the center of scientific discussion \cite{lipton2016mythos}.
A fully-interpretable model, is one whose decision-making process is completely transparent and predictable (e.g. a Decision Tree). However, this is virtually impossible to achieve when dealing with Machine Learning models, especially deep ones. For this reason we are forced to regress to a more lenient, but practical, definition of interpretability that coincides with \textit{Explainability}.

\begin{definition}
    A Machine Learning model is considered \textit{interpretable} if it can produce explanations for its decisions, that are understandable by humans.
    \label{def:interpretability}
\end{definition}

While the above definition can help impart the attribution ``interpretable" to a model, it can't help when comparing two models on their interpretability. In order to set the desired goal of enhancing a model's interpretability, the definition needs do be expanded to be able to quantify it.

To do this we will make the following claim: As a model's \textit{Interpretability} we will define its ability to designate which of its inputs it took into account when making a prediction (i.e. the \textit{contribution} of each input feature). 

\begin{example}
    Let the input of a sentiment analysis model be the sentence: \textit{``Although the food wasn't bad, we \underline{didn't} have a \underline{good} time."}
    
    An interpretable model, besides predicting that the sentence has a negative sentiment, should also provide an explanation. This can be accomplished by denoting that the words \textit{``didn't"} and \textit{``good"} had the largest influence in its final decision.

    \label{ex:sentiment}
\end{example}

The above claim allows us to form the final definition of a model's Interpretability.

\begin{definition}
    A model's Interpretability can be defined as the percentage of its input that it presents as the explanation.
    \label{def:interpretability_perc}
\end{definition}

According to Def. \ref{def:interpretability}, when a model can denote which of its input it took into account when making a prediction, it can be considered \textit{interpretable}. The percentage of its inputs that it denoted shows the amount of its Interpretability (Def. \ref{def:interpretability_perc}).

\begin{example}
    Following the same setting as in Example \ref{ex:sentiment}, $3$ models make predictions on the same sentence.

    \textbf{Model1}: ``\underline{Although} the food \underline{wasn't bad}, we \underline{didn't} have a \underline{good} time."
    
    \textbf{Model2}: ``Although the food \underline{wasn't bad}, we \underline{didn't} have a good time."
    
    \textbf{Model3}: ``Although the food wasn't bad, we \underline{didn't} have a \underline{good} time."

All $3$ models are Interpretable according to Def. \ref{def:interpretability}, as they all provide an explanation. The first model denotes all polarized words, the second all negatively polarized ones, while the third underlines only the most important words. This is why, out of these three models the third is the most interpretable (Def \ref{def:interpretability_perc}).
\label{ex:sentiment_2}
\end{example}

To quantify the amount of interpretability a model offers, the concept of an \textit{input mask} must first be defined.

\begin{definition}
    An input mask is a binary tensor, with the same shape as the input, whose purpose is to hide a portion of the input.
    \label{def:mask}
\end{definition}

By masking a portion of the input, less information is allowed to pass through to the model. If this information suffices for correct classification, then the input mask carries significant semantic value, as it shows the most important part of its input. This measure will be used to quantify a model's interpretability. If a model can specify which part of its input activated it the most when generating a prediction (e.g. \textit{which word(s) mattered the most for sentiment analysis}, or \textit{which area of the input image did it take into consideration when classifying the image}) then by Definition \ref{def:interpretability}, it is considered interpretable. Moreover, the percentage of the masked information, is directly related to the model's interpretability, as its reasoning becomes more specific (Def. \ref{def:interpretability_perc}). 

\begin{equation}
    \rm Interpretability({model}) = \mathnormal{I}(model) = \frac{masked \; features}{input \; features}
    \label{eq:interpretability}
\end{equation}


\begin{example}
    The Interpretability of each of the $3$ models from Example \ref{ex:sentiment_2} is:  
    
    \begin{align*}
    \rm Interpretability(model1) &= 6/11 \approx 54.5\%  \\
    \rm Interpretability(model2) &= 8/11 \approx 72.7\%  \\
    \rm Interpretability(model3) &= 9/11 \approx 81.8\%
    \end{align*}

\end{example}

First of all, the term \textit{Fidelity} is proportional to the \textit{performance} of a model at its task. This way it can be measured and compared. In the context of this study fidelity will be defined in a way to indicate how faithful a transparent network is to its black-box counterpart, i.e. how much of its original predictive power did it maintain. Thus it will be computed as the ratio of the interpretable network's performance to that of the original.

\begin{definition}
    The Fidelity of an Interpretable Machine Learning model is the ratio of its performance $P$ to that of its un-interpretable counterpart. 
    \label{def:fidelity}
\end{definition}

\begin{equation}
    \rm Fidelity({model}) = \mathnormal{F}(model) = \frac{\mathnormal{P}({model})}{\mathnormal{P}({baseline})}
    \label{eq:fidelity}
\end{equation}

A model's Fidelity shows the percentage of its initial performance that it managed to retain by becoming interpretable. Note that $P$ could be any metric that quantifies the model's performance at the given task.

\begin{example}
If a model achieves $60\%$ accuracy without the ability to explain its decisions and $50\%$ with that ability, then its Fidelity is $50/60 = 83.3\%$. Another way to think of this is that the model ``sacrificed" $17\%$ of its performance to become interpretable. Obviously, the higher the Fidelity, the more ``faithful" a model is to its initial performance.
\end{example}

Within this context, a couple of metrics can be defined that measure the aforementioned trade-off. 

\begin{definition}
    A model's \textit{Fidelity-to-Interpretability Ratio (FIR)} is defined as the model's Fidelity to the sum of its Fidelity and Interpretability.
    \label{def:fir}
\end{definition}

This metric essentially shows, how much of the model's interpretability was sacrifices for performance. It neither relays any information about the performance of the model, nor its interpretability; rather it shows the \textit{balance} between the two. To bound the ratio to $[0, 1]$ the following equation is used to compute it:

\begin{equation}
    \mathrm{FIR} = \frac{F}{F + I}
    \label{eq:fir}
\end{equation}

where $F$ is a performance metric (e.g. accuracy) and $I$ is the percentage of inputs masked. If the FIR is $1$ then the Fidelity of the model is much higher than its Interpretability. Likewise, when the FIR is close to $0$, the model's Interpretability severely outweighs its Fidelity. A balanced ratio of $0.5$ is ideal. 

Note that this score could be produced by underperforming on both categories. It cannot, therefore, be used to compare two models; for this purpose we will define a second metric. 

\begin{definition}
    The \textit{Fidelity-Interpretability Index (FII)} is defined as the product of a model's Fidelity with its Interpretability.
    \label{def:fii}
\end{definition}

\begin{equation}
    \mathrm{FII} = F \cdot I
    \label{eq:fii}
\end{equation}

With the help of the above definitions, we can formulate the goal of this study: \textit{the construction of a framework that is capable of producing interpretable Neural Networks with an arbitrary FIR} (i.e. the user can define the ratio of fidelity-to-interpretability). In essence we will attempt to create an architecture that hides the largest portion of its input, without losing its classification potency. Moreover, an extensive examination will be performed on how much fidelity is sacrificed for the sake of interpretability and vice versa (Sec. \ref{sec:fir}). 

While the framework introduced in this study is proposed as a general-use system, capable of working on any input data type (structured, text, image, etc.), the experimental part will be focused solely on image classifiers.

\section{Related Work}
\label{sec:related}

Due to the significant potential of Interpretable Neural Networks, a lot of research has been conducted on this topic \cite{bratko1997machine, ridgeway1998interpretable, nauck1999obtaining}. Some of the most popular techniques for visual interpretability rely on backpropagating the Network's prediction back through the network, to compute its derivative with respect to the input features (i.e. the pixels in the image). Through this, the contribution of each feature towards the final prediction is calculated \cite{zeiler2014visualizing, simonyan2013deep,  springenberg2014striving, ribeiro2016should, sundararajan2017axiomatic, lundberg2017unified, shrikumar2017learning}. These techniques, however, have been shown to have some major flaws, which deem them unfit for generating reliable explanations \cite{mahendran2016salient, ghorbani2019interpretation, kindermans2019reliability, nie2018theoretical, adebayo2018sanity}.

The most prominent example of interpretability in CNNs is Class Activation Mapping (CAM) \cite{zhou2016learning}. This technique involves concluding a CNN with a Global Averaging Pooling (GAP) and a single Fully Connected (FC) layer. This allows for the FC to learn a mapping directly from the high-level features extracted by the final convolutional layer to the classes. Thus, for any given image, it is possible to identify \textit{which extracted features activate each class the most}. By projecting these features onto the original image, a map of which area activates each class can be generated. This, in turn, allows a CNN to provide a degree of reasoning for its decision, i.e. \textit{where is it looking at when making a prediction}.
However, this comes at a cost of the model's complexity and therefore performance \cite{selvaraju2017grad, tagaris2019high},
essentially sacrificing the model's Fidelity for Interpretability.

The same principle has been applied in other studies as well, which tried to improve results by either replacing the GAP operation with Global Max Pooling \cite{oquab2015object} or a log-sum-exp pooling \cite{pinheiro2015image} operation. 
Another approach, called Grad-CAM \cite{selvaraju2017grad}, proposes an extension of CAM, where the importance of each feature map is determined not by the FC weights, but by the backward-propagating gradients. This allows for the use of a much broader range of architectures (e.g. no GAP required, more than one FC layers allowed), which lets networks achieve the same level of interpretability, without sacrificing performance. While these methods produce maps with continuous values, a binary mask (Def. \ref{def:mask}) can be generated via a simple threshold.

The core issue of CAM-based approaches is that they rely on the final convolution layer to generate the features that form the CAM. This layer, though, usually has a much smaller resolution than the input image and its CAM requires an image upscaling technique (e.g. bilinear interpolation) to be projected on the original image. This causes the CAM to be rather coarse (no fine details), thus reducing the interpretability of the model. This has been addressed in another study with the addition of an unsupervised network to improve the resolution and details of the CAMs \cite{tagaris2019high}, but to little effect.

Another family of techniques strive to achieve Interpretability by classifying perturbations of the input image. One way is to occlude different patches of the image and identify which occlusions result in a lower classification score (i.e. when relevant objects are occluded the model will have a hard time to classify the image) \cite{zeiler2014visualizing}. Another study attempts to classify many patches containing a pixel then average these patch class-wise scores to provide the pixel’s class-wise score \cite{oquab2014learning}. These approaches are computationally inferior to the CAM-based methods, as they require multiple passes of an input image. However, at the cost of higher computational complexity, they can theoretically achieve an arbitrarily-high level of Interpretability, while maintaining model Fidelity; in practice, however, high levels of interpretability are impractical. Interestingly, the first approach is conceptually similar to ours, however, instead of relying on heuristically hiding parts of the image, we propose a fully trainable process, offering a higher degree of both Fidelity and Interpretability at a much lower computational cost.

The proposed framework will consist of two networks, one tasked at producing input masks and an image classifier. The two will be trained jointly, in a collaborative manner, on two objectives: to minimize the classification loss and maximize the number of masked pixels. The joint model will learn to hide as much of the image as possible, while retaining a high classification performance.


The main contributions of this study can be summarized as follows:

\begin{itemize}

    \item A novel framework for discriminative localization in classification tasks, is proposed. Instead of relying on coarse CAMs or computationally-expensive image perturbations, our framework is fully trainable, producing the finest masks possible in a single inference. 
    
    \item The above framework, also has the ability of manual FIR (Def. \ref{def:fir}) designation, i.e. the user can decide how fine or coarse his masks will be (fine masks can come at the cost of performance). This way the user can have an active participation in where the model stands regarding the fidelity-interpretability trade-off.
    
    \item Given this choice the model is guaranteed to have the highest level of fidelity (i.e. highest possible classification performance) or interpretability (i.e. finest masks possible).
    
    \item An exploration is conducted on the cost of interpretability to a model. I.e. \textit{How much of its performance does it need to sacrifice to be interpretable?}
    

\end{itemize}

\section{Theoretical background}

One main component of the framework 
is the production of the binary mask (Def. \ref{def:mask}). To achieve this, the continuous output of the previous layer must be converted to binary. This paper explores two ways of performing this conversion: a deterministic approach and a stochastic one.

\subsection{Binary Neuron formulation}

The mathematical formulation for both types of binary neurons will be presented here.

\subsubsection{Deterministic thresholding}
\label{sec:deterministic}

The most intuitive way of converting a continuous value to a binary one is to select a threshold and set the value equal to $1$ or $0$ if it is above or below the threshold. Given a continuous variable $z \in \mathbb{R}$ and a threshold $\tau$, this is equivalent to a Heaviside step function $H$ offset by $\tau$:

\begin{equation} 
H(z) = \mathbbm{1}_{z \geq \tau} =
\begin{cases}
1, & \text{if $z \geq \tau$}\\
0, & \text{if $z < \tau$} \\
\end{cases}
\label{eq:step}
\end{equation}

To effectively select the threshold $\tau$, the continuous values need to be normalized to a predetermined range (e.g. $[0, 1]$). Practically, this can be achieved by applying a sigmoid function to $z$.

\begin{equation}
\sigma(z) = \frac{1}{1 + e^{-z}}
\label{eq:sigm}
\end{equation}

Though the sigmoid function can easily be added after any neural network layer, this is not the case for the step function (Eq. \ref{eq:step}).
From its definition, the derivative of the step function is $0$ everywhere besides $\tau$, where $H$ is not differentiable.

\begin{equation}
    \frac{dH(z)}{dz} = 0 \;\; \forall z \neq \tau
\end{equation}

Normally, this would prevent the error from back-propagating to layers preceding the step function, which would make their training impossible. A typical workaround is to ignore this layer during backpropagation (i.e. treat it as an identity function $f(z) = z$, which has a derivative of $1$). 

\begin{equation}
    \frac{dH(z)}{dz} \approx 1
    \label{eq:deterministic_approcimation}
\end{equation}

The name \textit{deterministic thresholding} attempts to distinguish this approach from the \textit{stochastic} one that will be described in Section \ref{sec:stochastic}.

\subsubsection{Binary Deterministic Neurons}
\label{sec:bdn}

By combining the aforementioned ideas, a type of new type of neuron can be defined, the \textit{Binary Deterministic Neuron} (BDN). This includes an affine transformation, a non-linear activation function (i.e. sigmoid) and a deterministic thresholding function:

\begin{equation}
H(x) = H \left( \sigma(h(x)) \right) = H \left( \sigma \left( W x + b \right) \right)
\label{eq:bdn}
\end{equation}

where $W$ and $b$ are trainable parameters and $h(x) = W x + b$. The derivatives of this layer with respect to its parameters $W$ and $b$ are:

\begin{align}
\begin{split}
\frac{\partial H}{\partial W} 
&= \frac{\partial H}{\partial \sigma} \cdot \frac{\partial \sigma}{\partial h} \cdot \frac{\partial h}{\partial W} \\
&\approx 1 \cdot \sigma(h)  \cdot (1-\sigma(h)) \cdot \frac{\partial (W x + b)}{\partial W} \\
&= \sigma(h)  \cdot (1-\sigma(h)) \cdot x
\end{split}
\end{align}

and 

\begin{align}
\begin{split}
\frac{\partial H}{\partial b} 
&= \frac{\partial H}{\partial \sigma} \cdot \frac{\partial \sigma}{\partial h} \cdot \frac{\partial h}{\partial b} \\
&\approx 1 \cdot  \sigma(h)  \cdot (1-\sigma(h)) \cdot \frac{\partial (W x + b)}{\partial b} \\
&= \sigma(h)  (\cdot 1-\sigma(h))
\end{split}
\end{align}

This is only an approximation of the gradient, as the derivative of the step function was substituted with $1$ (Eq. \ref{eq:deterministic_approcimation}). Another way to think about this would be as having an infinitely steep sigmoid function (i.e. one that resembles a step function).

The same principle can be applied to convolutional layers simply by substituting the affine transformation with another with its equivalent. 

\subsubsection{Stochastic thresholding}
\label{sec:stochastic}

Another thought would be to convert a continuous variable to binary through a \textit{stochastic thresholding} function. The motivation for this approach arose from the various visual attention models, which saw better results with stochastically generated attention masks (what they refer to as ``hard attention") rather than deterministic ones (i.e. ``soft attention") \cite{mnih2014recurrent, ba2014multiple, xu2015show}.

Contrary to the step function $H$ (Eq. \ref{eq:step}) which is deterministic (i.e. if $z$ is greater than the threshold, $H(z)$ will always be equal to $1$), stochastic functions carry with them a degree on uncertainty. Here, there isn't a threshold to compare $z$ to; instead $z$ becomes $0$ or $1$ at random, with a probability equal to its value. Obviously, $z$ needs to be normalized (through Eq. \ref{eq:sigm}) for this to work properly.

\begin{equation}
B(z) = \mathbbm{1}_{P(z)} =
\begin{cases}
1, & \text{with probability $z$}\\
0, & \text{otherwise}
\end{cases}
\label{eq:stoch}
\end{equation}

This means that if $z = 0.7$, $B(z)$ will be equal to $1$ with a probability of $70\%$. In this sense, $B(z)$ is essentially a Bernoulli random variable with $p=z$. 

Computing the derivative of this function is impossible, even by making the approximation of Equation \ref{eq:deterministic_approcimation}. The problem, in this case, is the noise induced by the stochasticity of the operation, which will need to be taken into account in the approximation.

\subsubsection{Binary Stochastic Neurons}

Similar to a BDN (introduced in Sec. \ref{sec:bdn}), a \textit{Binary Stochatic Neuron} (BSN) performs an affine transformation, a non-linear activation function and a stochastic threshold:

\begin{equation}
B(x) = B \left( \sigma(h(x)) \right) = B \left( \sigma \left( W x + b \right) \right)
\label{eq:binary_sigmoid}
\end{equation}

Backpropagating through this neuron is trickier, due to its stochastic nature. Since a derivative can't be computed the traditional ways, it has to be estimated. 

\subsection{Gradient estimation}

The simplest way of formulating the problem of gradient estimation is to consider an architecture where the noise is not inherently part of the units but is injected into them. 
This way the output of each BSN is a function (i.e. the step function in Eq. \ref{eq:step}) of both a noise source $n$ and the result of a transformation on its inputs (in this case $\sigma(h(x)) = \sigma(W x + b)$).

\begin{equation}
B(x) = H (\sigma(h(x)) +  n)
\end{equation}

This decomposition of the neuron's function creates a path for the gradients to flow during backpropagation; the only problem is the estimation of the gradient despite the noise $n$. Four separate gradient estimators will be examined for this purpose. All of these will be viewed through the lens of a single BSN, $B$, but with no loss of generality this can be applied to any and every BSN in the network. 

\subsubsection{Likelihood-ratio estimation}

The most general technique for estimating the gradient through stochastic units is the \textit{Likelihood-ratio method} \cite{glynn1990likelihood} also known as the \textit{Score Function estimator} \cite{kleijnen1996optimization} or the \textit{REINFORCE estimator} \cite{williams1992simple}. 

The goal of a non-stochastic network is to minimize a cost function $J$ w.r.t the trainable parameters of the network $\theta$ 
(i.e. $W, b \in \theta$). This can be done through gradient descent, which requires the computation of the gradient $\nabla_{\theta} J(\theta)$.

Due to the existence of one or more noise sources $N$ in the network (which directly affect the value of the cost function), the goal in stochastic networks is to minimize the expected value of the cost over the noise sources w.r.t. $\theta$, through its gradient $g$.

\begin{equation}
g = \nabla_{\theta} \E_N \left[ J(\theta) \right]
\end{equation}

The computation of $g$ is infeasible and has to be estimated. 
Through the likelihood-ratio method, the unbiased estimator $\hat{g_u}$ can be derived. This takes the following form.

\begin{equation}
\hat{g_u} = J \cdot (H - \sigma(h))
\label{eq:uncentered_reinforce}
\end{equation}

The subscript $u$ is added to denote that this is the \textit{uncentered} form for the estimator. A proof for the equation above is provided by \cite{bengio2013estimating}. 
This estimator is convenient as it only requires broadcasting $J$ throughout the network (i.e. no need for back-propagation), which makes training the network possible. 

\subsubsection{Variance reduction}

Even though the above estimator is unbiased, it has quite a high variance. There are several methods that help reduce this, the most prominent is by subtracting a \textit{baseline} $\tilde J$ from the cost. The baseline that leads to the greatest reduction in variance for the estimator is:

\begin{equation}
\tilde J = \frac{\E[(H - \sigma(h))^2 \cdot J]}{\E[(H - \sigma(h))^2]}
\label{eq:reinforce_baseline}
\end{equation}

Out of all possible baselines this results in lowest variance, while not increasing the estimator's bias. The proof of this can also be found in \cite{bengio2013estimating}.
Because $H$ and $h$ are specific to a single neuron, the baseline can be though of as a weighted average of the cost values $J$, whose weights are specific to that particular neuron \cite{bengio2013estimating}.
By subtracting Eq. (\ref{eq:reinforce_baseline}) from the cost $J$ in Eq. (\ref{eq:uncentered_reinforce}), the \textit{centered} estimator $\hat{g_c}$ is derived:

\begin{equation}
  \hat{g_c} = (J - \tilde J) \cdot (H - \sigma(h))
  \label{eq:centered_reinforce}
\end{equation}

\subsection{Straight-Through estimators}

Another idea would be to completely ignore the step function $H$ and the noise induced by the stochasticity of the neuron. This is referred to as the \textit{Straight-Through estimator} (ST). 
There are two variants of this idea, depending on whether or not the derivative of the sigmoid is ignored during computation. The two estimates of the derivative of $J$ w.r.t the weights $W$ would be

\begin{equation}
    \hat g_{st}^{(1)}(W) = \frac{\partial \sigma(h(W))}{\partial W} = \sigma(h) \cdot (1 - \sigma(h)) \cdot x
\label{eq:straight_through_estimator_1}
\end{equation}

or

\begin{equation}
    \hat g_{st}^{(2)}(W) = \frac{\partial h(W)}{\partial W} = x
\label{eq:straight_through_estimator_2}
\end{equation}

These two estimators are very simple to compute, but they are clearly biased as $\E [\hat g_{st}(W)] \neq \partial \E [J] / \partial W$. The reason for this bias is the discrepancy between the functions in the forward and backward pass.

\subsection{Slope-Annealing trick}
\label{sec:sa}

Another approach is called the \textit{Slope-Annealing trick} \cite{DBLP:journals/corr/ChungAB16} and attempts to reduce the bias of the \textit{Straight-Through estimator}. This draws inspiration from the fact that the sigmoid function becomes steeper if multiplied by a scalar larger than $1$. As this scalar increases, the sigmoid approaches the step function $H$ (Eq. \ref{eq:step}), while remaining continuous and differentiable. Following the formulation of Equation \ref{eq:binary_sigmoid}:

\begin{equation}
B_{sa}(x) = B \left(s \cdot \sigma(h(x)) \right) = B \left(s \cdot  \sigma \left( W x + b \right) \right)
\label{eq:slope_annealing_forward}
\end{equation}

where $s$ is the aforementioned scalar that will be referred to as the ``slope". By applying the first \textit{Straight-Through} estimator (Eq. \ref{eq:straight_through_estimator_1}) to the slope-augmented sigmoid, the estimator for this layer is derived:

\begin{equation}
    \hat g_{sa}^{(1)}(W) = \frac{\partial \sigma(h(W))}{\partial W} = s \cdot \sigma(h) \cdot (1 - \sigma(h)) \cdot x 
\label{eq:slope_annealed_estimator}
\end{equation}

The trick is to start $s$ from a value of $1$ and slightly increase it during training, so that the sigmoid resembles more and more a step-function \cite{DBLP:journals/corr/ChungAB16}.  

\section{Proposed Framework}
\label{sec:hns}

\begin{figure*}[ht]
    \centering
    \includegraphics[width=450 pt]{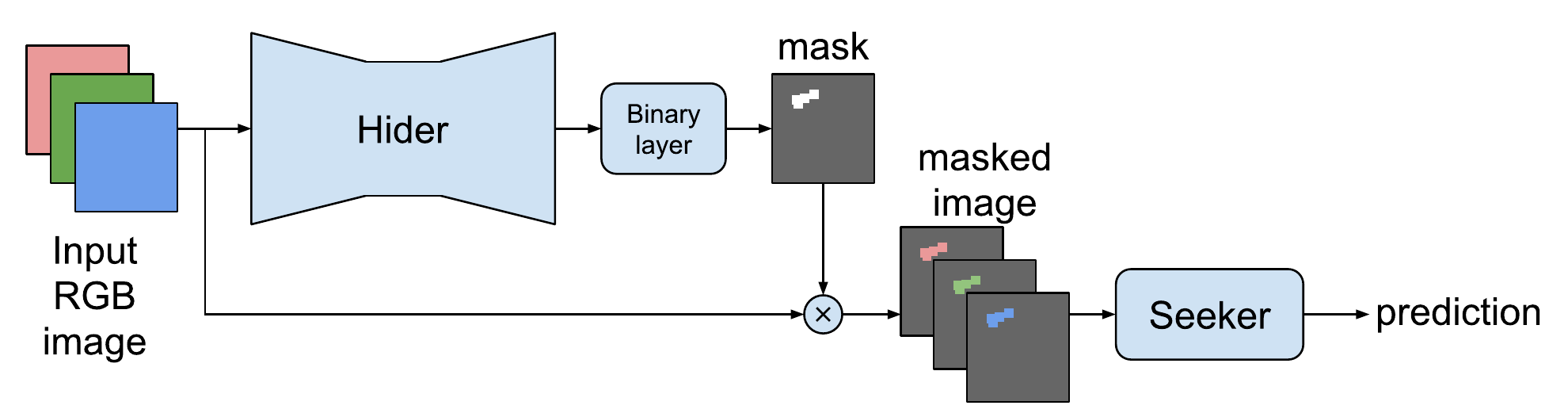}
    \caption{Hide-and-Seek architecture for image classification.}
    \label{fig:hns}
\end{figure*}

The driving idea behind this paper is to train a Deep Learning architecture for classification, with the ability to provide a certain ``reasoning" for its decision. This reasoning comes by method of discriminative localization, i.e. the ability to highlight a region in the input image that the network mainly took into account to make its decision. The goal is to produce a network that can be trained to both classify an image \textit{and} produce a mask that indicates the part of that image that the network used to produce its prediction.

The architecture should have the ability of generating the finest mask possible (i.e. highest degree of interpretability) for a given classification performance (i.e. level of fidelity). By sacrificing some of this performance, it is possible to achieve even higher degrees of interpretability. This allows for an active selection of where the framework stands on the fidelity-interpretability trade-off. 


\subsection{High-level Architecture outline}

As stated previously, the proposed framework consists of two networks trained jointly. The first, which is tasked at generating the binary masks, is a model whose output must have the same dimensions as its input. Because its role is to ``hide" a portion of its input, this network will be referred to as the \textit{hider}. The second, is any regular classifier that can work well on its own in the aforementioned task and will be referred from now on as the \textit{seeker}.

Thus, the only limitation lies in the design of the hider. If the input is structured (e.g. a table), the output will also be a table of the same shape. If the input data is a sequence (e.g. text) the output should be a sequence of the same length. Finally, as is our case, if the input is an image, the output should also be an image with the same resolution. There are many architectures which can be used to generate the binary masks, for example an Autoencoder or even an image segmentation network like a U-Net \cite{ronneberger2015u}. 

The goal of the hider, in our case, is to produce a binary mask for the input image that will hide some of its pixels. It is trained so that it leaves the pixels with the most relevance to classification, while hiding the background. Additionally, it will be rewarded for hiding as many pixels as possible from the input image, in hopes of producing concise masks. However, it will not be trained directly, but through the use of the seeker, an idea inspired by the Generative Adversarial Networks \cite{goodfellow2014generative}. 

The seeker, on the other hand, is trained to classify the masked images produced by the hider, In contrast to the hider, the seeker will have direct access to the loss function and will serve the goal of back-propagating the error back to the hider.

The \textit{Hide-and-Seek (HnS)} framework, essentially is the collaborative training of a hider and a seeker. 
The architecture of a HnS model for image classification is depicted in Figure \ref{fig:hns}. The input image is fed to the hider, which with the help of a binary layer produces a binary mask. This mask is then applied to the input and fed to the seeker, which attempts to generate its prediction.

This methodology could be applied to different tasks besides image classification. For example by swapping the hider with a sequence-to-sequence model and the seeker with a sequence classifier, this could be applied to Natural Language Processing problems such as sentiment analysis. It wouldn't be hard to imagine a model capable of masking the irrelevant words in a sentence. 

The loss function will have two terms, a standard classification error and a function of the number of pixels in the hider's mask. The first term serves to train the models for classification, while the second encourages the hider to produce smaller masks. These two terms are to an extent competing one another, as smaller masks might make the images harder to classify. However, if they are balanced, the trained model will be capable of producing the smallest masks that achieve a sufficiently high performance in classification.

\subsection{Hider}

\begin{figure*}[ht]
    \centering
    \includegraphics[width=450 pt]{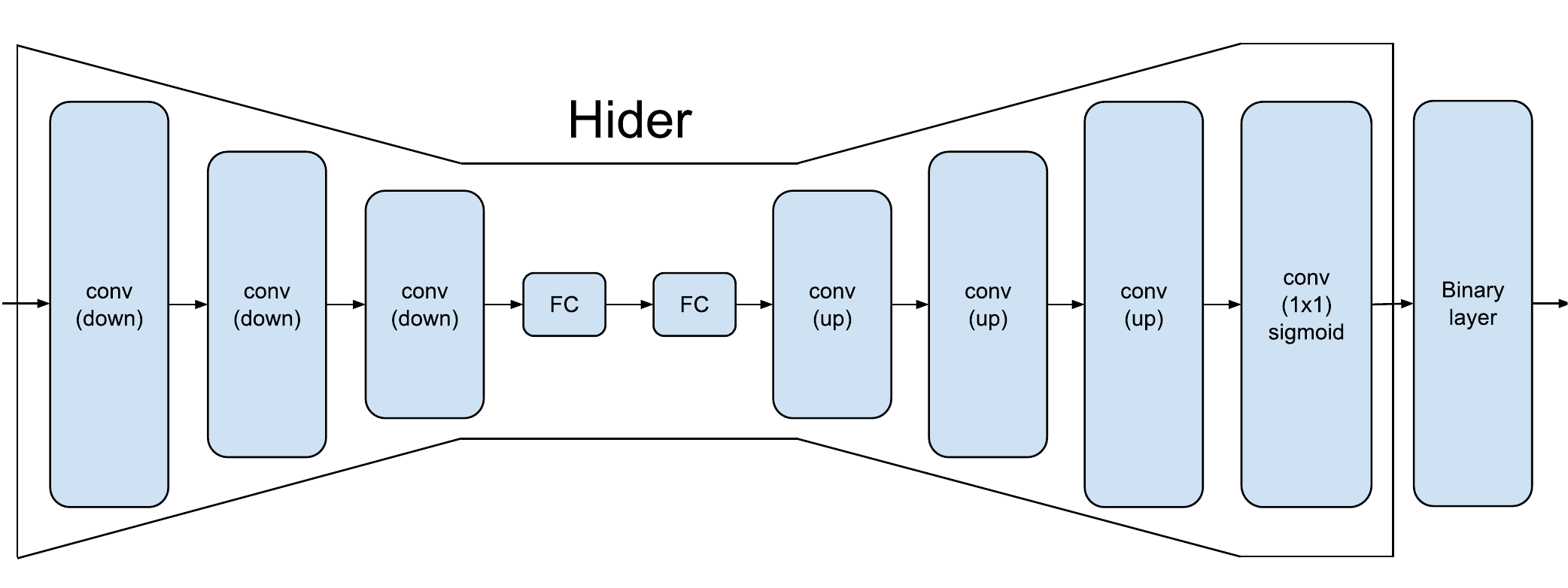}
    \caption{Convolutional Autoencoder as a hider.}
    \label{fig:hider}
\end{figure*} 

Though there are many architectures which could be used effectively for generating a binary mask, a Convolutional Autoencoder was elected for this study, due to its simplicity. Intuitively, the hider should learn to recognize what parts of the image might be important and what might not, for classification. This should have sufficient capacity in its downscaling path to extract the necessary features for classification. Depending on the size of the input images and the difficulty of the task, the complexity of the hider should also be adjusted (e.g. larger networks are needed for more difficult tasks).  

Figure \ref{fig:hider}, outlines a template architecture, which can be modified with more or less convolution layers to accommodate tasks of varying difficulty. The final two layers cannot be changed. The final hider layer is a $1 \times 1$ convolution layer, which is meant to produce a mask with a single channel that outputs values within $[0, 1]$. Its output is fed to a binary layer, which converts its real valued inputs to binary either through deterministic (Sec. \ref{sec:deterministic}) or stochastic thresholding (Sec. \ref{sec:stochastic}). As a convention, it is not considered to be part of the hider.

Two architectures were employed in the present study both of which follow the one of Figure \ref{fig:hider}. The first one, features $7$ layers ($2$ downscale conv, $2$ FC, $2$ upscale conv, $1$ final conv) and $585,677$ trainable parameters and was used on the ``smaller" datasets. The second included $13$ layers ($5$ downscale conv, $2$ FC, $5$ upscale conv, $1$ final conv) and $12,910,635$ parameters. This was used on the ``large" dataset.

\subsection{Seeker}

The seeker, is a regular CNN, that could be used to perform image classification on the desired dataset, without any modifications. Its input should have exactly the shape as the hider's, while outputting the probability that the input belongs to each class. There are no limitations regarding its capacity, though 
it is recommended to be lower or equal to that of the contraction path of the hider.

Three different seeker architectures were examined: A small CNN with $4$ layers ($3$ conv, $1$ FC) and $185,066$ trainable parameters, a larger CNN with 6 layers ($5$ conv, $1$ FC) and $5,407,342$ trainable parameters, and finally a ResNet-50 ($50$ layers, $52,931,854$ total parameters) \cite{he2016deep}.

\subsection{Loss Function}

In order to train the joint network for both classifying images and masking their content as best as possible, a loss function with two terms needed to be constructed.

The first term represents the classification loss $J_{\mathrm{clf}}$ between the prediction and the actual target (in our case cross-entropy). This is required to train the network for classification.

The second term  ($J_{\mathrm{mask}}$) was a measure of the amount of information that the mask allows to pass through it. While a few different metrics were considered, such as the percentage of pixels equal to $1$, or the energy of the masked image, even something as simple as the sum of the pixels of the mask worked without issue.

The weighted sum of these two terms was used as the joint network's loss function.

\begin{align}
    \begin{split}
        J &= \alpha \cdot J_{\mathrm{clf}} + (1 - \alpha) \cdot J_{\mathrm{mask}} \\
        &= \sum_{i=1}^N{\left[\alpha \cdot y_i \log{\hat y_i} + (1 - \alpha) \cdot  \sum_{j=1}^M H_{i,j}\right]}
        \label{eq:loss}
    \end{split}
\end{align}

The hyper-parameter $\alpha \in [0, 1]$ regulates the amount each of the two losses contributes to the joint loss function.

\subsubsection{Relation of the loss regulator $\alpha$ to the training objective}
\label{sec:relation}

From Equation \ref{eq:loss} it becomes apparent that a value of $\alpha$ near $1$ leads to the joint loss being dominated by the classification loss. Empirically, this encourages the hider to not mask anything from the original image, as the classifier can use the extra information for improving its performance (which is more important to the total loss for large values of $\alpha$).

On the other hand, values near $0$ encourage the hider to mask the whole image, as the classification loss becomes irrelevant.

Through this parameter, the user gains control over the FIR (Def. \ref{def:fir}),
as higher values of $\alpha$ lead to an increased model fidelity, while lower values lead to an increased interpretability. However, to train a model that is actually useful, a balance between the two needs to be established. The goal set in this present study was to hide as much information as possible, without impacting the seeker's performance.

During training, in some occasions the hider converged to a sub-optimal solution where it masked the exact same pixels for each image and couldn't update further. This issue arose more frequently for smaller values of $\alpha$ and only for deterministic thresholding.

\subsubsection{Adaptive weighting}
\label{sec:adaptive}

In order to achieve the aforementioned goal and overcome the issue mentioned previously, a scheme was devised for dynamically adapting $\alpha$ during training.

More specifically, the joint network started its training with $\alpha=1$ (i.e. pure classification). The classification loss was monitored during training. If the loss stagnated for a few iterations then the value of $\alpha$ was dropped by a small amount, causing the mask loss' importance to increase. This, encourages the hider to hide more pixels, which, in turn, make classification harder for the seeker, leading to a temporary destabilization of the classification loss. After a while it will stabilize and stagnate again; when this happens, $\alpha$ is further decreased. This process is repeated until a very small value of $\alpha$ is achieved.

This technique alleviates the need for selecting a proper value for $\alpha$ beforehand, or the need of grid-searching over this hyper-parameter. Additionally, it stabilizes the training process, because it emphasizes on training the seeker early on, while the hider comes into play during the final iterations. In fact, no instances of the issue mentioned in Section \ref{sec:relation} were observed when using an adaptive $\alpha$. Furthermore, a decrease in convergence time was observed when using an adaptive $\alpha$. This speedup is speculated to be caused by the progressive training of the seeker in evermore masked images. Finally, this technique also serves as an early-stopping mechanism, as the loss is monitored and gives an indication on when to stop training.

Even through this technique an issue occasionally arose, where the hider would adopt extreme strategies, either masking everything or nothing at all. This means that the total loss was determined by only one of its two terms. These instances will be referred to as \textit{collapses}, referencing the ``mode collapse" of GANs. \cite{goodfellow2014generative}

\subsection{Implementation Details}

Three possible ways of speeding up the training process were examined: pre-training the hider, the seeker or both and will be discussed in detail in Section \ref{sec:pretraining}.

The pre-training of the hider was accomplished by training it, in an unsupervised manner, like an Autoencoder: the inputs and targets are set as to be the same, while the hider is trained on a reconstruction loss \cite{goodfellow2016deep}.
In the case of RGB input images, the target should be the same image converted to grayscale to make in compatible with the single-channel bianry masks. 
The seeker is much easier to pre-train, as it can be accomplished by training it like a regular CNN.

To implement the previously-mentioned adaptive $\alpha$ scheme, a queue of the past $100$ classification losses is kept. If none of these diverge significantly from the average loss of the queue then the value of $\alpha$ will be decreased. In this case, the model's weights are stored and running average is flushed. The condition used was that the loss values shouldn't fluctuate more than $10\%$ of the average loss.

\section{Experiments}

\subsection{Datasets}
Experiments were performed on three datasets:

\begin{itemize}
    \item the \textit{MNIST}\footnote{http://yann.lecun.com/exdb/mnist/}
    dataset, which consists of $60,000$, $28 \times 28$ grayscale images from handwritten digits (10 classes in total).
    \item the \textit{Fashion-MNIST}\footnote{https://github.com/zalandoresearch/fashion-mnist}
    dataset, which consists of $70,000$, $28 \times 28$ grayscale images from fashion products (10 classes in total).
    \item the \textit{CIFAR10} and \textit{CIFAR100}\footnote{https://www.cs.toronto.edu/~kriz/cifar.html}
    datasets, which consists of $60,000$, $32 \times 32$ RBG images distributed amongst $10$ and $100$ classes respectively.
\end{itemize}


Most of the experiments and conclusions were made on the CIFAR10 experiments, because its smaller size allowed for a larger number of experiments.

\subsection{Evaluation criteria}
\label{sec:criteria}

There are two criteria that the models can be evaluated on: \textit{performance} and \textit{variance}. 

Performance has two components, \textit{classification performance} (i.e. how accurately the seeker classifies) and \textit{masking performance} (i.e. what percentage of pixels does the hider mask), both of which are important. 
The first can be measured by the model's Fidelity (Eq. \ref{eq:fidelity}), while the second through its Interpretability (Eq. \ref{eq:interpretability}).

As discussed in Section \ref{sec:definition},
there is a trade-off between the two, which can be regulated through the $\alpha$ parameter (Sec. \ref{sec:relation}).
To properly measure this balance two additional metrics are used: the FIR (Eq. \ref{eq:fir}) and the FII (Eq. \ref{eq:fii}).

\begin{figure}[h]
    \centering
    \includegraphics[width=\linewidth]{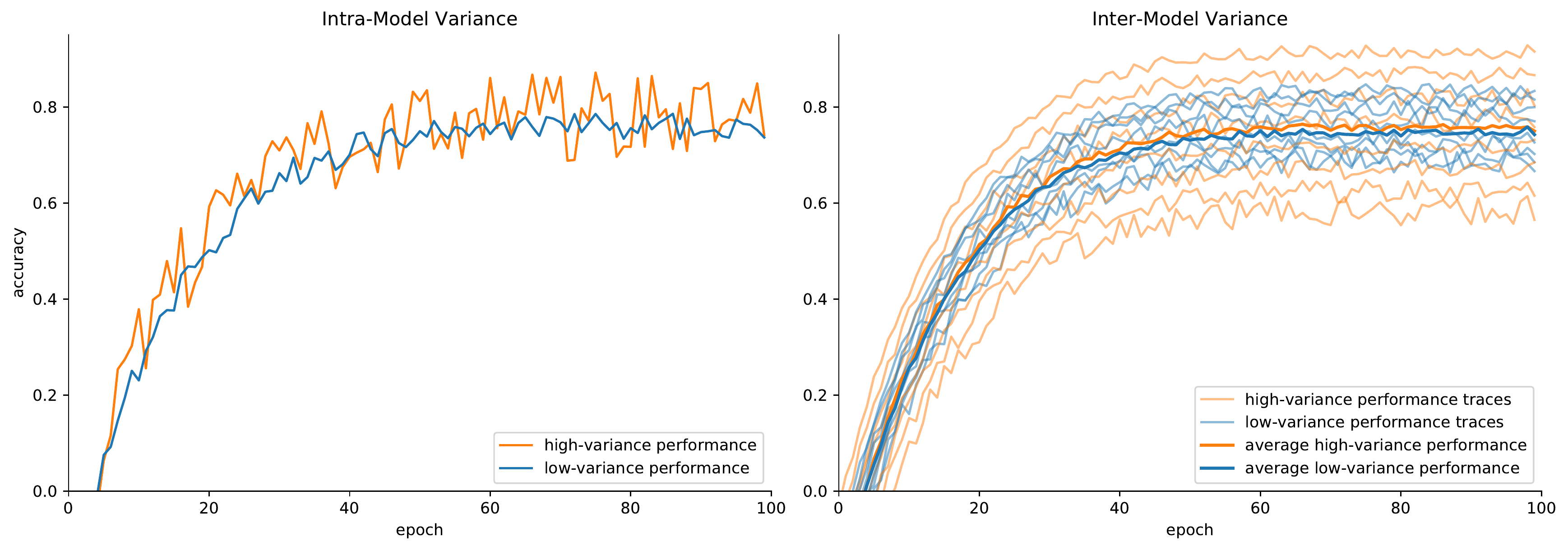}
    \caption{Representation of the two forms of variance exhibited in a model. In both vases the average performance is the same, but the variance is higher in the orange models. The intra-model variance (left) appears as fluctuations in the training curve. The inter-model variance (right) can be seen through the training curves of $16$ models with similar intra-model variances.}
    \label{fig:hns_variance}
\end{figure}

Likewise, there are two forms of variance that the models can exhibit: \textit{intra-model variance} (i.e. deviations in performance within the same model from epoch to epoch) and \textit{inter-model variance} (i.e. deviations in performance from model to model). The first can be identified through fluctuations in the model's training curves, while the second requires re-training the same model for a number of times. These two forms of variance are depicted in Figure \ref{fig:hns_variance}.

For the experiments conducted, the models were trained for a total of $10$ times, under the same conditions, to be able to detect the latter form of variance. For simplicity, from now on both types will be referred to as \textit{variance}. Note that while variance is linked to performance, it doesn't require high-levels of it, rather consistency. A model can have a low variance while performing poorly if it its performance doesn't fluctuate from epoch to epoch and from training to training. 

\subsection{Full training vs Pretraining}
\label{sec:pretraining}

The first question that arose, was whether or not the hider and seeker required any form of pretraining in order for the HnS model to converge. For this reason, the same model was trained on the cifar10 dataset, with four different initialization conditions (i.e. training from scratch, pretrained hider, pretrained seeker, both hider and seeker pretrained). As mentioned in Section \ref{sec:criteria}, each of these models was trained $10$ times independently to properly assess the model's variance.

\begin{figure}[h]
    \centering
    \includegraphics[width=0.6\linewidth]{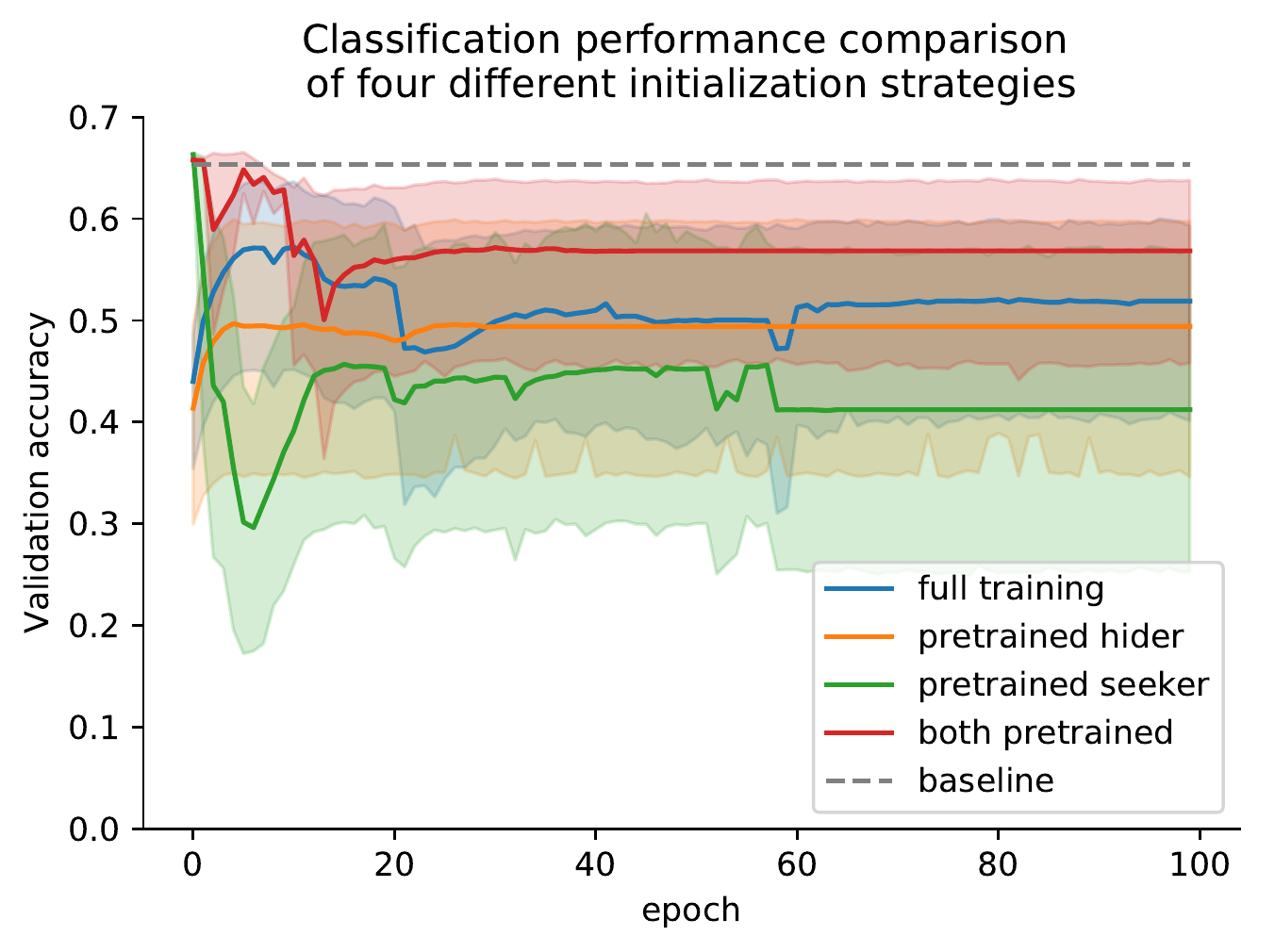}
    \caption{Classification performance for a deterministic HnS trained on the CIFAR10 dataset, under different initialization conditions.}
    \label{fig:init_class}
\end{figure}

The first thing to check is the classification performance of the models. Figure \ref{fig:init_class}, illustrates the performance of each of the four initialization conditions. The grey line is the baseline performance which was achieved by a fully trained seeker on the same dataset. By hiding portions of the image, it is expected that the performance will experience a slight dropoff, which can be attributed to the fidelity-interpretability tradeoff (see Sec. \ref{sec:definition}). The bold colored lines represent the mean performance of the $10$ models of each initialization, while the shade gives an indication of the inter-model variance. 

The models with both components pretrained proved to be the best and most stable in this category. Interestingly, training a model initialized from a pretrained seeker, proved to be the toughest of all, as its former training didn't translate well when the hider started masking pixels.

Due to the goal set in Section \ref{sec:definition}, the aim was to select models that don't experience a steep dropoff in terms of classification performance. For this reason, only models that perform within $90\%$ of the baseline were examined further. A secondary task was to hide as much information as possible. For this reason an arbitrary threshold of $90\%$ of pixels hidden was established. If these two goals are achieved, then the model will be considered to have converged to an ``optimal" solution. A total of $4$ models with a pretrained hider fulfilled these requirements. Only $1$ from the rest each of the rest categories managed to do so.

This shows that by having a pretrained hider, it is much easier to train a HnS model and achieve optimality. It should be mentioned that a lot of the models models that did not achieve this status, still managed to converge to sub-optimal solutions. 

\begin{figure}[h]
    \centering
    \includegraphics[width=0.6\linewidth]{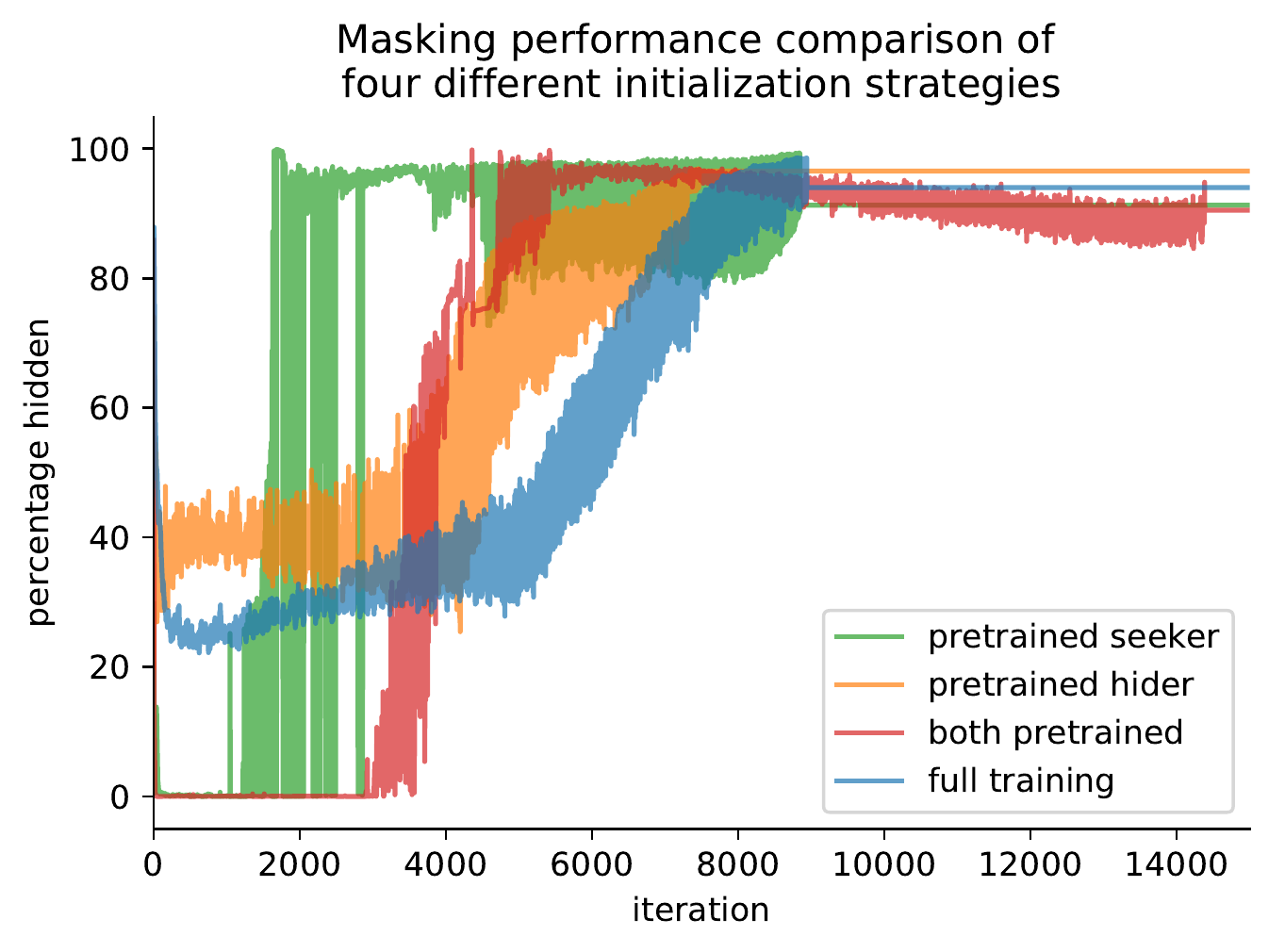}
    \caption{The best model's masking performance under different initialization conditions.}
    \label{fig:init_mask}
\end{figure}

The masking performance of the best model of each initialization condition is depicted in Figure \ref{fig:init_mask}. The two models with a pretrained seeker experienced much sharper ascends, while the rest were much more stable during training. Especially the one with just the pretrained seeker, experienced a lot of variance. This gives some indication as to why so many of these models collapsed. The pretrained hider has an interesting training curve, it starts at a rather high percentage (due to the pretraining of its component) but drops a bit as training proceeds. The reason is that, in order to increase its classification performance (which is dominant during the early stages of training due to the adaptive weighting), it allows more information to pass. When it achieves a low enough classification error, it proceeds again to hide more and more pixels and was actually the first of the four to converge.

\begin{figure}[h]
    \centering
    \includegraphics[width=0.6\linewidth]{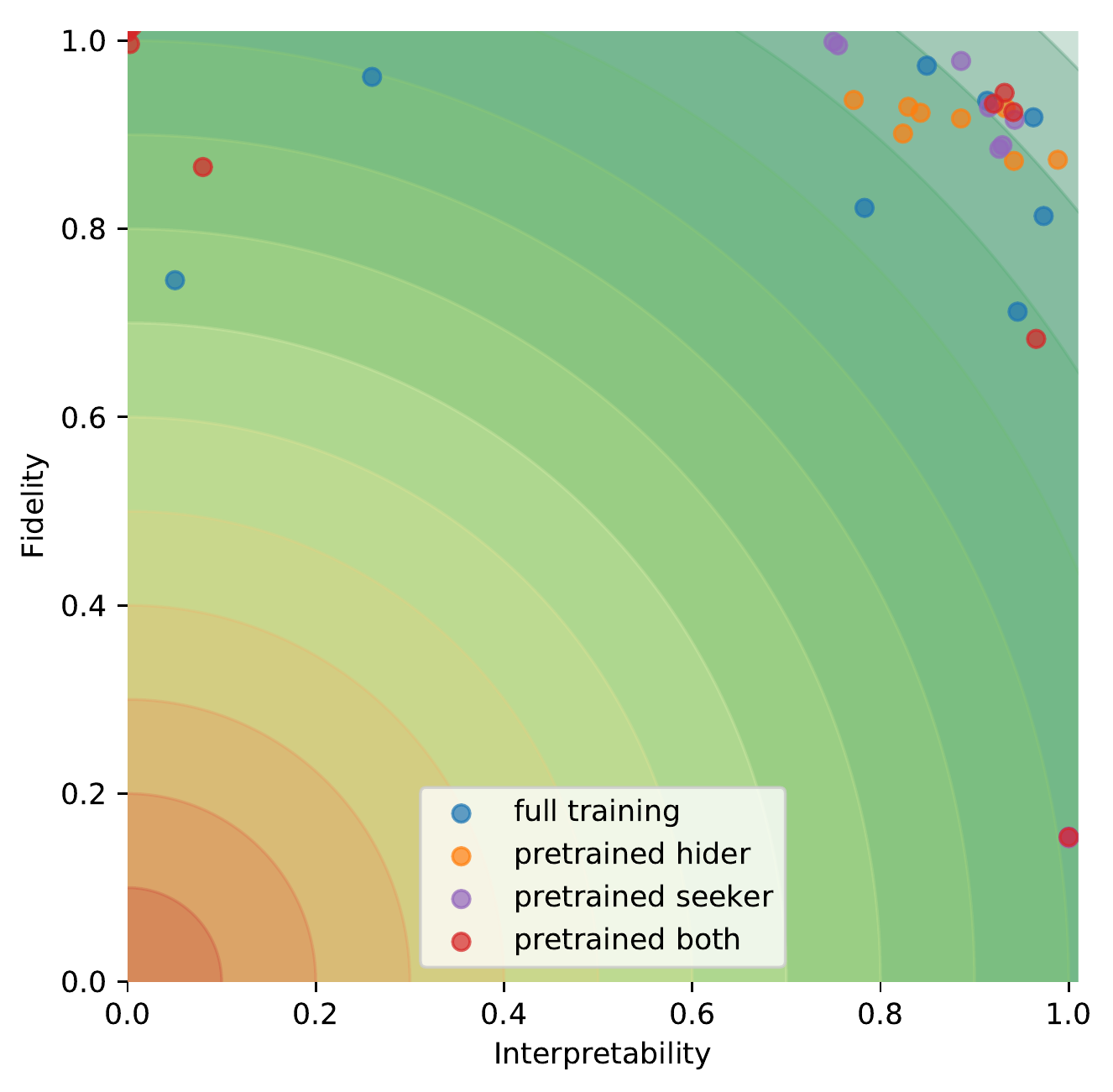}
    \caption{Fidelity-Interpretability projection for different initialization strategies.}
    \label{fig:init_2d}
\end{figure}

Perhaps the best way to assess performance in both of these areas is to project the Fidelity and Interpretability of all the models on 2 axes. Figure \ref{fig:init_2d} depicts this projection.

This paints a clearer picture regarding the different strategies. While the pretrained hider didn't lead to any collapses, it did settle to many suboptimal solutions. Instead by having both components pretrained, led to a lot of optimal convergences, even though it suffered a few collapses.




\subsection{Stochastic estimator comparison}

Another objective of this research was to examine which of the stochastic estimators performs the best. Four different estimators were examined: Straight-Through v1 (ST1), Straight-Through v2 (ST2), Slope-Annealing (SA) and REINFORCE. Like before, each model was trained $10$ times independently and was initialized from scratch.

\begin{figure}[h]
    \centering
    \includegraphics[width=0.6\linewidth]{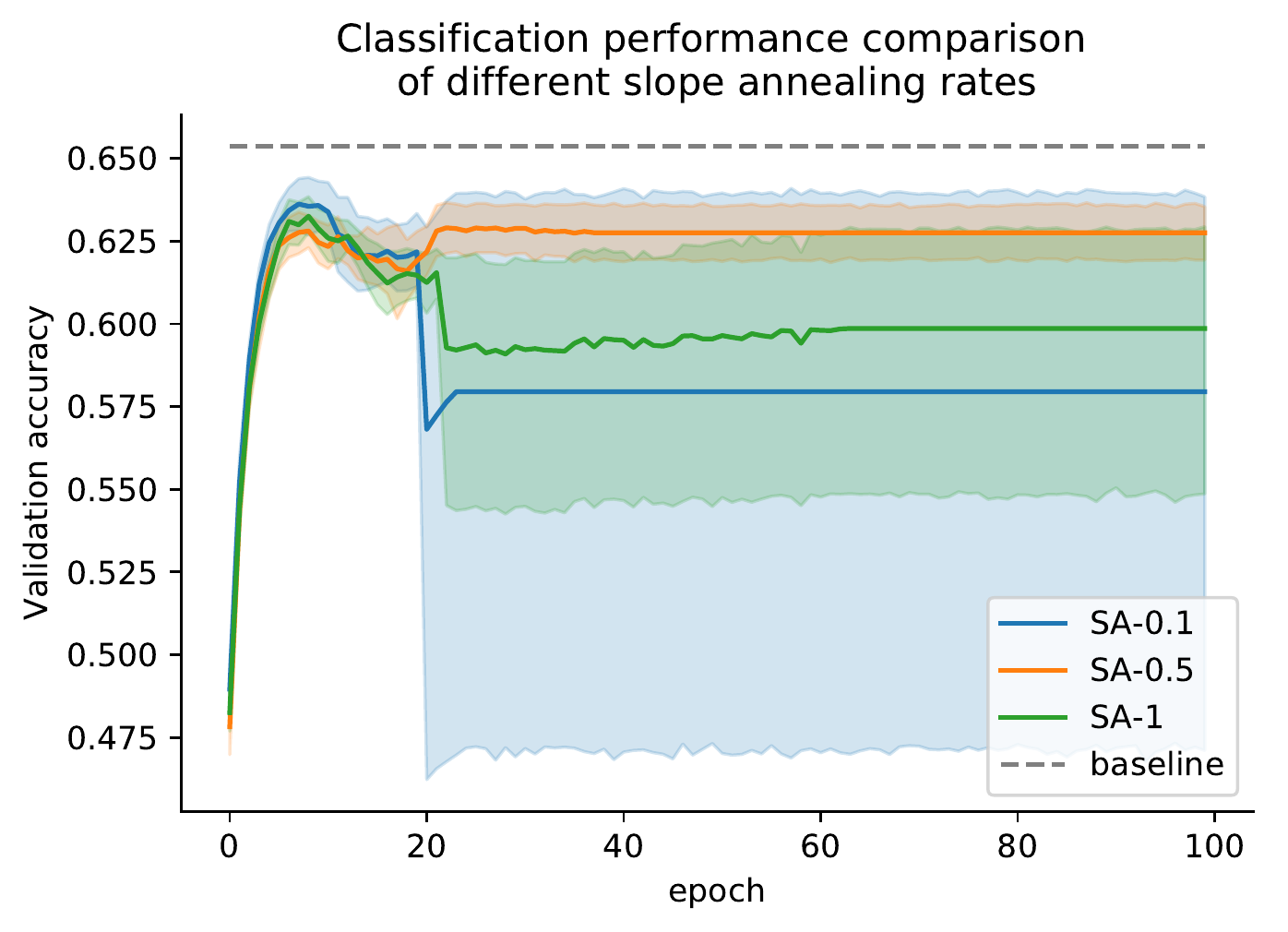}
    \caption{Classification performance of models trained with three different slope-annealing rates.}
    \label{fig:sa_class}
\end{figure}

In the case of the SA estimator, different rates of slope increase were examined. When applying this estimator, the slope $s$ starts off at $1$ and increases gradually during training (Sec. \ref{sec:sa}). A small increase was applied at the end of each weight update. The rate of the slope, indicates the percentage of its increase at the end of each epoch. For example, a rate of $0.5$ means that the slope increases by a total of $50\%$ at the end of each epoch. First of all, the best rate needed to be identified. Five different rates were examined, including $0.1$, $0.5$, $1$, $10$ and $100$. Only the middle three had models that converged to optimal solutions, namely $2$, $2$ and $1$. Figure \ref{fig:sa_class} shows the classification performance of these three rates. This Figure, along with the previous numbers, give an indication that indication that smaller rates, which lead to more gradual increases of the slope are better. The models with a rate of $0.5$, clearly achieve better performance, while exhibiting a lower overall variance.

\begin{figure}[h]
    \centering
    \includegraphics[width=0.6\linewidth]{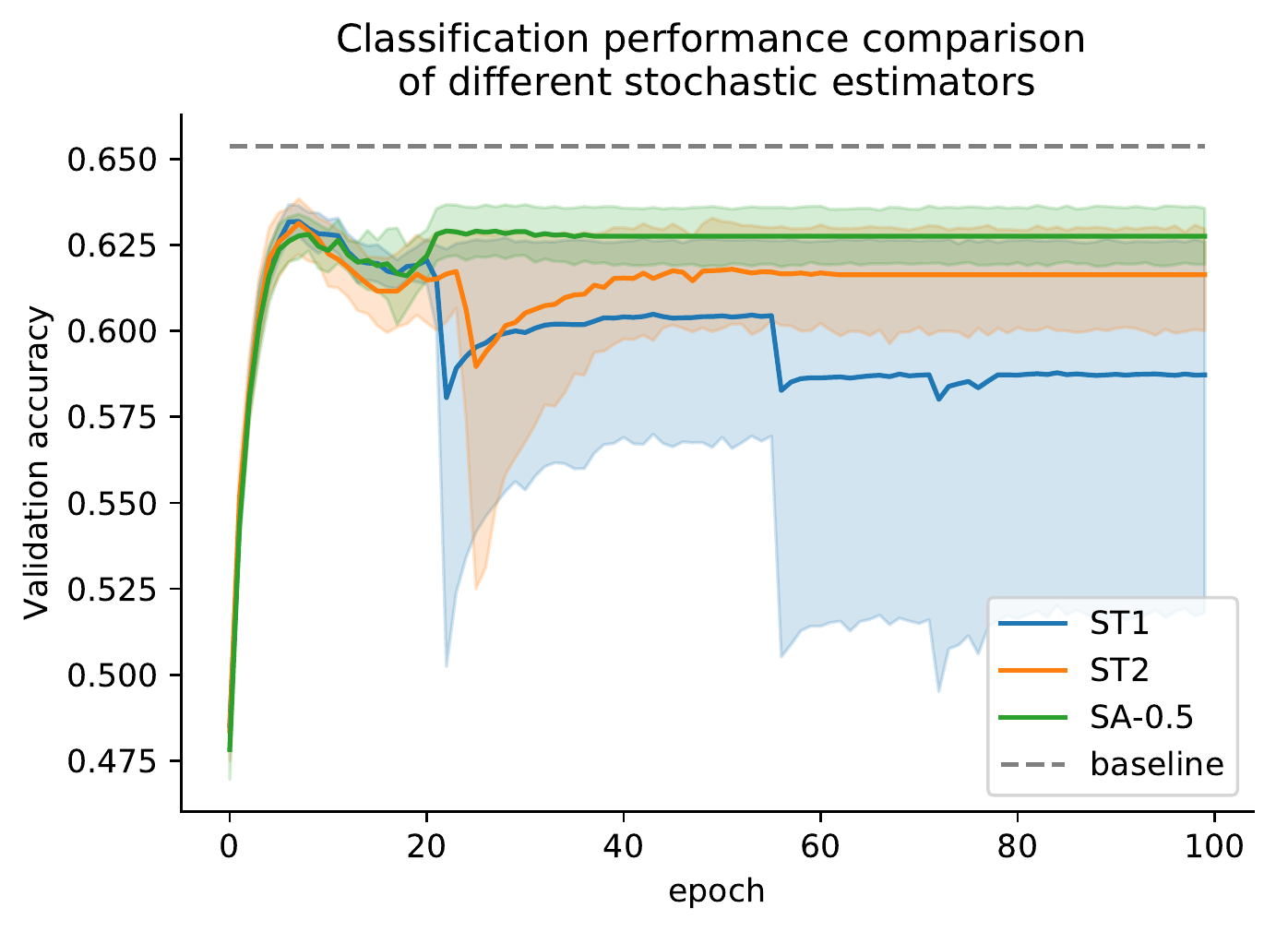}
    \caption{Classification performance of stochastic models trained with three different gradient estimators.}
    \label{fig:stoc_class}
\end{figure}

The next step was to compare the different types of estimators. Out of the four, ST1 had $4$ optimal convergences, ST2 just had $1$, SA (Rate=$0.5$) had $2$ and REINFORCE $0$. The last, while being the best known gradient estimator couldn't be made to work for this problem. 
The classification performance of these estimators can be seen in Figure \ref{fig:stoc_class}. SA seems to be ahead in terms of the rest, with ST2 following closely behind. 

\begin{figure}[h]
    \centering
    \includegraphics[width=0.6\linewidth]{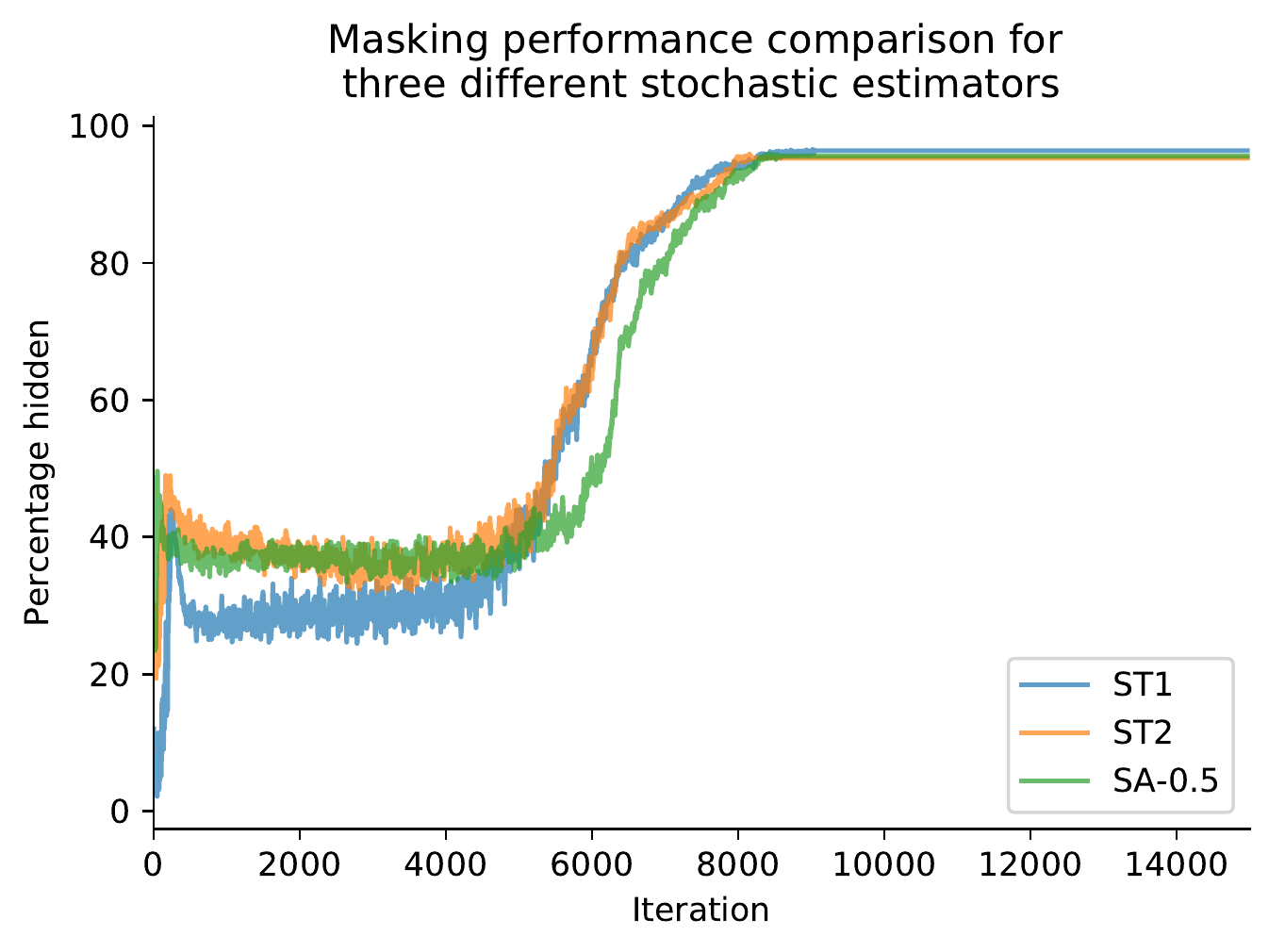}
    \caption{Masking performance of stochastic models trained with three different gradient estimators.}
    \label{fig:stoc_mask}
\end{figure}

The masking performance of the best model from each estimator is portrayed in Figure \ref{fig:stoc_mask}. While all three models converged to the same percentage, the two ST estimators managed to get there slightly faster. Interestingly, despite the stochasticity in the training process, these estimators don't exhibit a lot of variance.

\begin{figure}[h]
    \centering
    \includegraphics[width=0.6\linewidth]{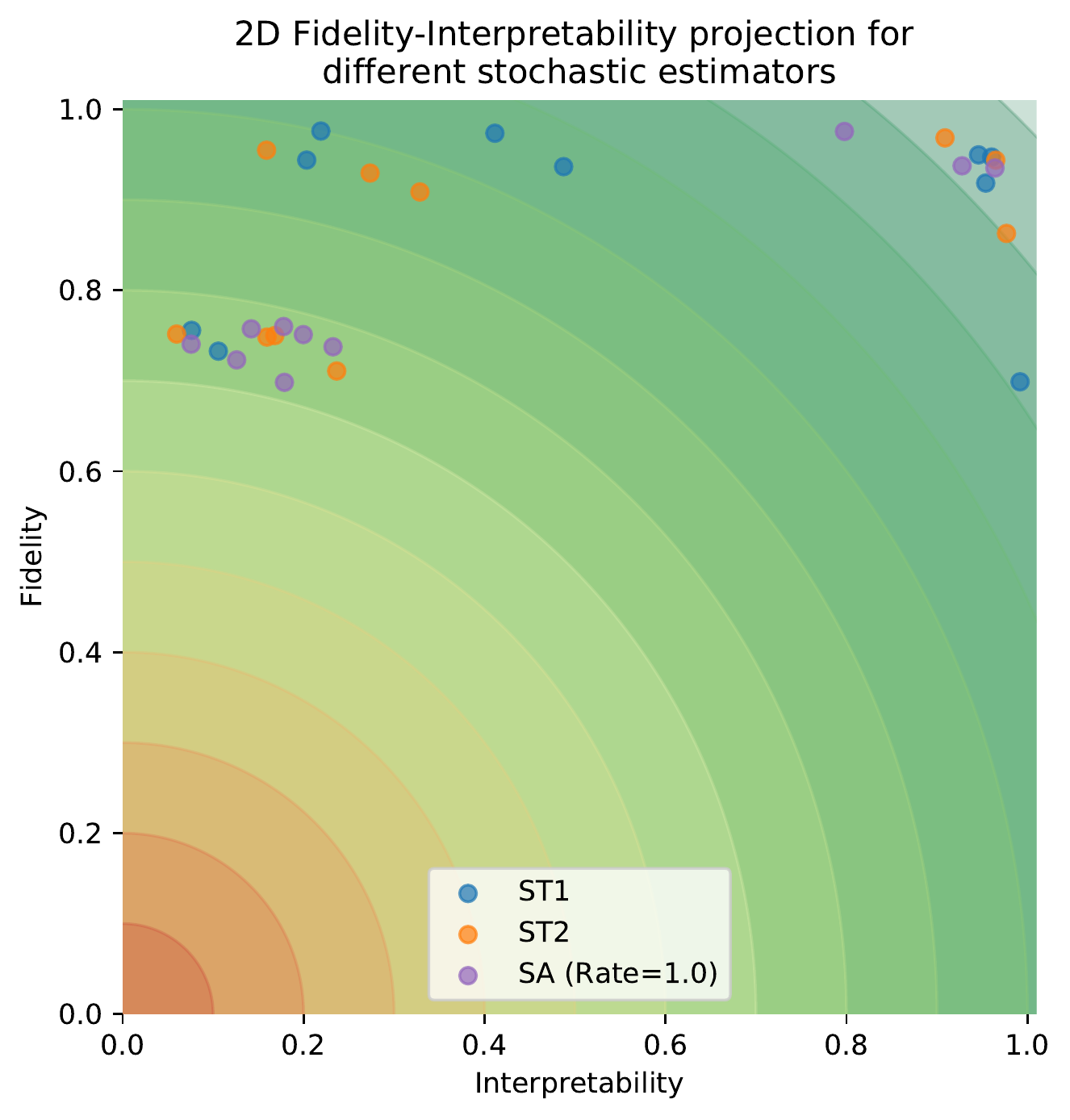}
    \caption{2D Fidelity-Interpretability projection of stochastic models' performance.}
    \label{fig:stoc_2d}
\end{figure}

These two performances are known to come with a tradeoff, which cannot properly be assessed by just looking at the best model's performance. To visualize both of the terms that come into the FII equation (i.e. Fidelity and Interpretability), a 2D projection of the models' performance performance could be made (Fig. \ref{fig:stoc_2d}). In this the Interpretability of a model is depicted in the x-axis, while its Fidelity is depicted in the y-axis. The performances of the models in each run are scattered as dots throughout the graph. Note that solutions near left are considered as collapsed, due to their low Interpretability. Same thing for solutions near the bottom, regarding Fidelity. Optimal solutions can be found near the top right.  

All of these models achieve a high degree of Fidelity. In fact none of the models managed to collapse in their classification performance. Interpretability, on the other hand, seems to vary a lot from model to model. Only a handful of models managed to converge to optimal solutions.

A large number of models seems to be stuck in with $[0.7, 0.8]$ Interpretability and $[0.1, 0.3]$ Fidelity. It is unclear why this area is so common in collapses, or if these models would ever manage to get themselves out of this region, if trained for more epochs. One theory is that by hiding the wrong pixels (i.e. crucial ones towards classification, the model suffers a great loss in Fidelity ($\approx 20-30\%$), without managing to significantly increase its Interpretability.

Concerning this plot, its hard to assess the best stochastic estimator, however ST1 seems a bit more stable than the rest.

\subsection{Deterministic vs Stochastic}
\label{sec:det_vs_stoch}

One of the main objectives was to determine if the stochastic thresholding would yield superior results to the deterministic. The straight-through estimator (ST1) proved to be, marginally, the best from all stochastic estimators and will be used to represent the ``stochastic" family. For a fair comparison the deterministic models that were trained from scratch will be used to represent the ``deterministic" family. 

The models were evaluated on a basis of performance (both on the scales of Fidelity and Interpretability) as well as variance, numbers of collapsed models and convergence speed.

The quantitative results are presented in Table \ref{tab:comparison}.

\begin{table}[h]
    \centering
    \begin{tabular}{c | c | c}
                                                      &       Det.    &        Stoch.      \\ \hline \hline
        \#models collapsed ($<20\%$ Fidelity)         &        1      &     \textbf{0}     \\
        \#models collapsed ($<20\%$ Interpretability) &  \textbf{2}   &     \textbf{2}     \\
        \#models collapsed (total)                    &        3      &     \textbf{2}     \\ \hline
        
        mean convergence Fidelity $^1$                &      0.86     &    \textbf{0.88}   \\ 
        peak convergence Fidelity $^1$                &      0.96     &    \textbf{0.98}   \\ \hline

        mean convergence Interpretability $^1$        & \textbf{0.80} &         0.65       \\ 
        peak convergence Interpretability $^1$        &      0.92     & \textbf{1.00} $^2$ \\ \hline
        
        mean FII $^1$(Eq. \ref{eq:fii})             &      0.79     &   \textbf{0.82}    \\ \hline
        peak FII $^1$(Eq. \ref{eq:fii})             &      0.85     &   \textbf{0.91}    \\ \hline
        
        mean FIR $^1$(Eq. \ref{eq:fii})             & \textbf{0.53} &        0.60        \\ \hline

        average convergence (epochs) $^1$             &      38.3     &    \textbf{31.6}   \\
        fastest convergence (epochs) $^1$             &  \textbf{19}  &         20         \\

    \end{tabular}
    \caption{Table comparing stochastic to deterministic thresholding}
    \label{tab:comparison}
\end{table}

\begin{figure}[h]
    \centering
    \includegraphics[width=0.6\linewidth]{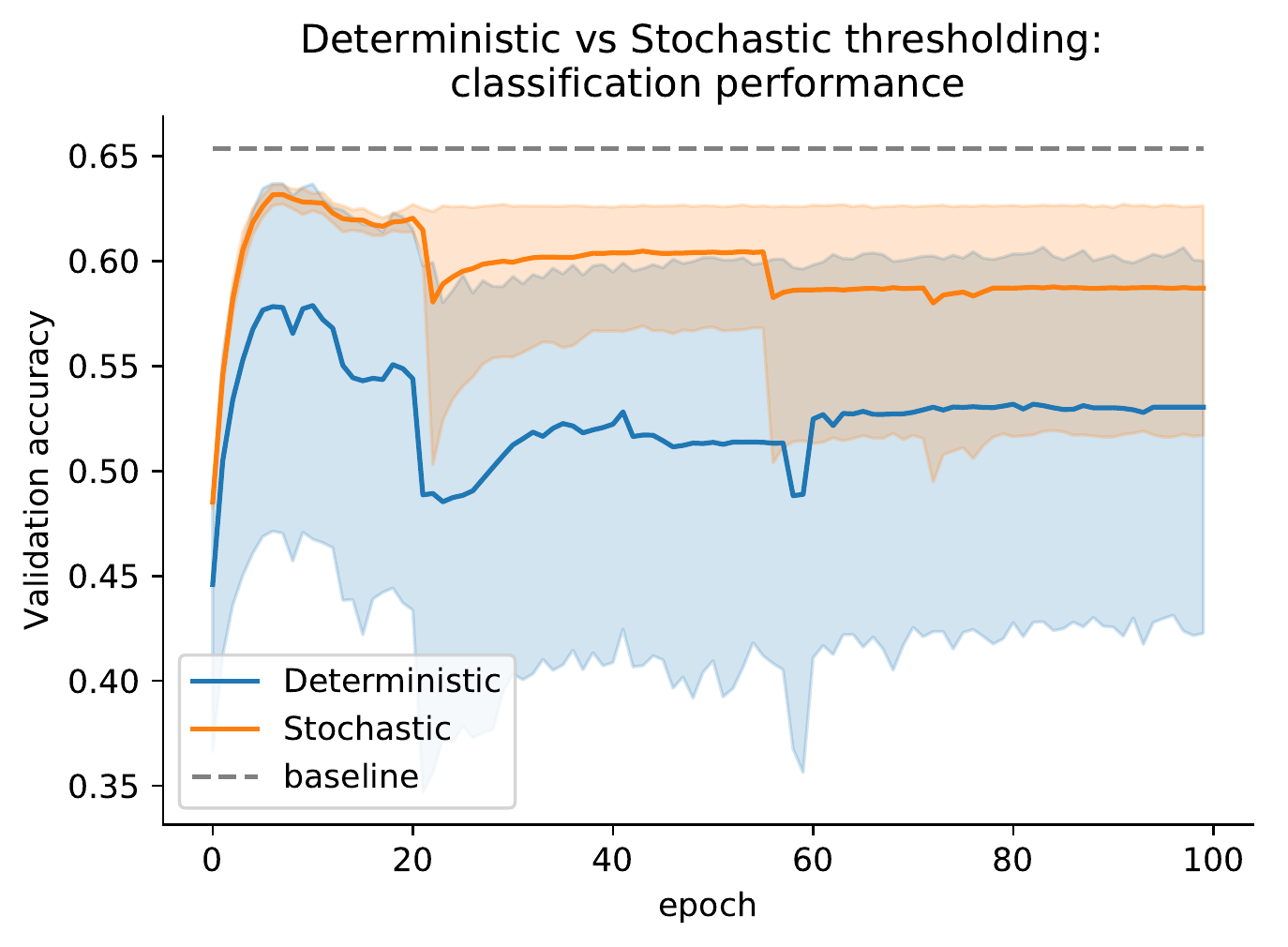}
    \caption{Classification performance of deterministic and stochastic models.}
    \label{fig:det_stoch_class}
\end{figure}

First, regarding their classification performance (i.e. Fidelity), stochastic models seem to be on top. Granted, the difference in terms of peak and mean performance isn't much, however, all other evidence clearly favors stochastic models. None of the stochastic models collapsed (compared to $1$ deterministic), while exhibiting a lower variance and were generally much more stable, as indicated by Figure \ref{fig:det_stoch_class}.

\footnotetext[1]{excluding collapsed models}
\footnotetext[2]{this model actually had an Interpretability of $0.997766$, with a fidelity of $0.429047$; obviously, a perfect Interpretability score would mean that the model masks the entire image and is predicting at random (which would cause a collapse).}

\begin{figure}[h]
    \centering
    \includegraphics[width=0.6\linewidth]{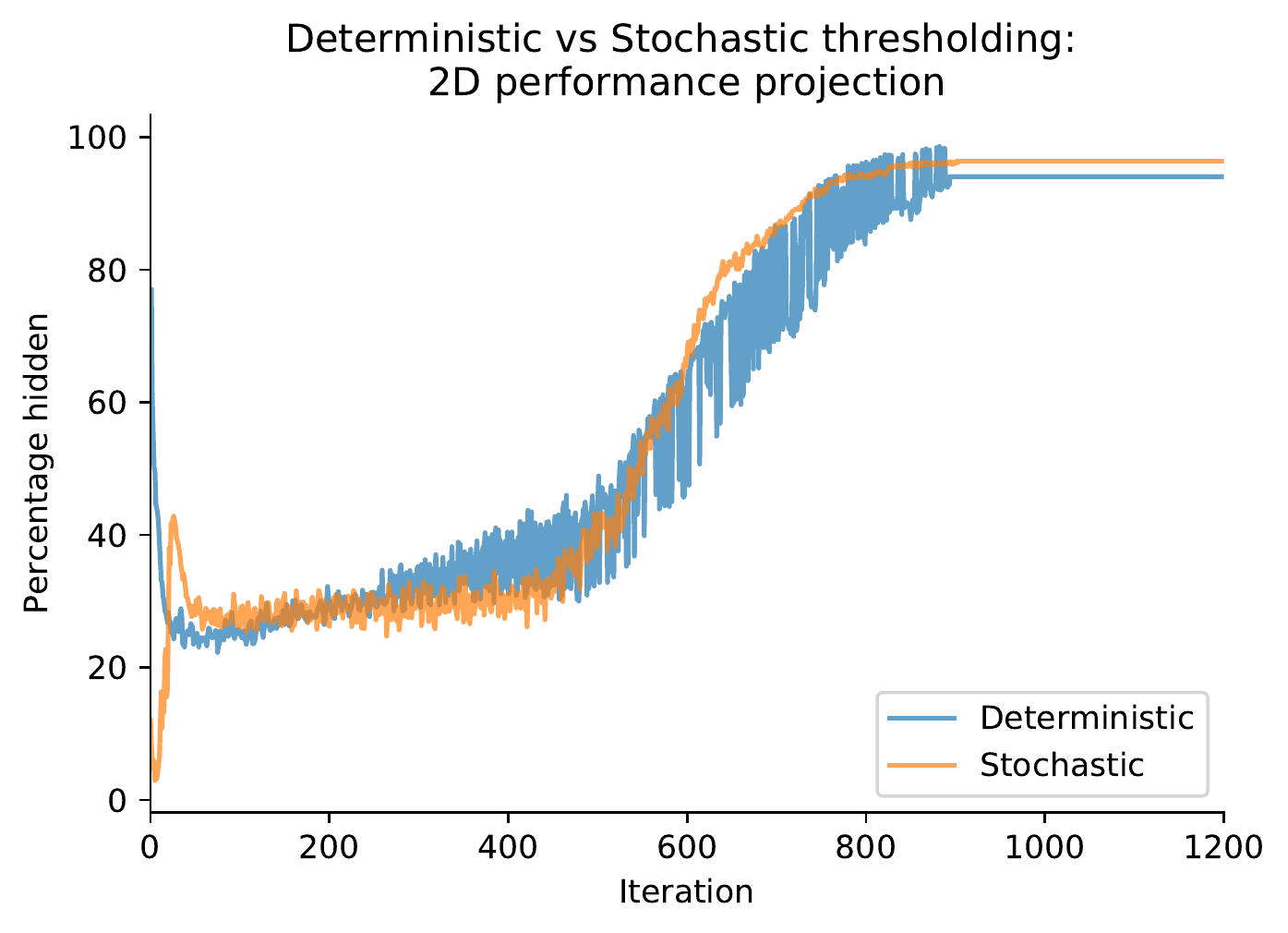}
    \caption{Masking performance of best deterministic and stochastic models.}
    \label{fig:det_stoch_mask}
\end{figure}

Masking, on the other hand seems to favor deterministic models in average, however the best overall masking model was a stochastic one. Stochastic models tend to have a lot of inter-model variance, causing inconsistency from experiment to experiment. On the other hand, These models don't have a lot of intra-model variance, because of their stochastic nature for generating masks. Not a much can be said in terms of collapses, as both models had the same number. The percentage hidden of the best performing deterministic and stochastic models can me seen in Figure \ref{fig:det_stoch_mask}. Out of these two the stochastic reached a higher percentage, while being a lot more stable and a bit faster. While this seems to be the case for these specific models the trend is that deterministic estimators are the best.

Both of these observations can be confirmed by the models' FIR scores (Eq. \ref{eq:fir}). Stochastic models seem to favor classification over masking, thus leading to an $FIR > 0.5$. Deterministic models are a bit more balanced, in this regard (i.e. $FIR \approx 0.5$), which results in a better and more consistent masking performance. A more in-depth analysis will be offered in Section \ref{sec:fir}.



\begin{figure}[h]
    \centering
    \includegraphics[width=0.6\linewidth]{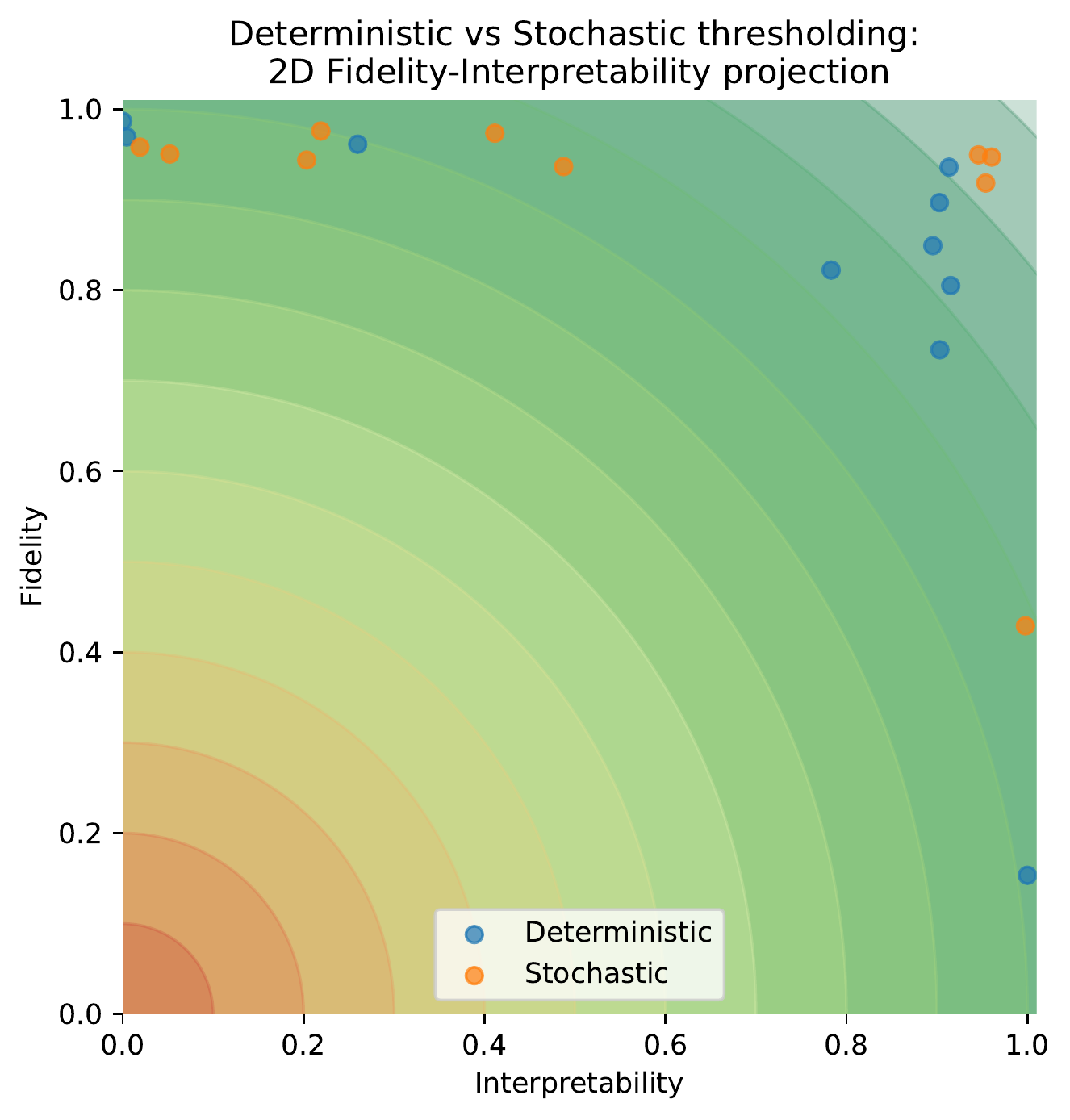}
    \caption{Fidelity and Interpretability of deterministic and stochastic models projected in 2D.}
    \label{fig:det_stoch_2d}
\end{figure}

Another way to inspect the tradeoff for these two families of models is to project their performance on two axes, Interpretability and Fidelity (Fig. \ref{fig:det_stoch_2d}).


This figure portrays the contrast of the two different types of thresholding. While deterministic models seem to favor masking over classifying, in stochastic models the opposite is happening. They seem to ignore the percentage of pixels hidden, while always striving for high classification performance. This is one of the reasons for their reduced variance, as will be discussed in Section \ref{sec:discussion}.

\subsection{Results}

It should be noted that the aim of this research is not to push the boundaries on classification performance, but to augment existing classifiers with Interpretability. For this reason, all models that were tested on the HnS framework, were first trained on their own, without any masking to establish a baseline. This baseline is used to compute the model's \textit{Fidelity} (Eq. \ref{eq:fidelity}).

The best performing models for each of the three datasets are presented in Table \ref{tab:results}.

\begin{table}[h]
    \centering
    \begin{tabular}{c | c | c | c | c | c | c}
                       &   Estimator       &  Epoch  &  Fidelity (\%)  & Interpretability (\%) &    FIR     &  FII  \\ \hline
         MNIST         & Deterministic     &    8    &     $98.83$     &        $98.46$        &   $0.501$    & $0.973$ \\
         Fashion-MNIST & Deterministic     &    7    &     $99.54$     &        $98.67$        &   $0.502$    & $0.982$ \\
         CIFAR10       & SA (rate = $0.1$) &   24    &     $95.95$     &        $96.15$        &   $0.499$    & $0.922$ \\
         CIFAR100      & Deterministic     &   35    &     $92.46$     &        $95.51$        &   $0.492$    & $0.883$ \\
         
    \end{tabular}
    \caption{Best performing model for each dataset.}
    \label{tab:results}
\end{table}

The first two datasets, were used to push both Fidelity and Interpretability as high as possible, as they are significantly easier than the rest. All types of models and estimators performed well on these datasets, while no collapses were exhibited at all. On average models managed to obtain both a Fidelity and Interpretability of $98\%$. All models converged to exactly $0.5$ FIR during training.

Around one out of three models converged to near-optimal solutions in the CIFAR10 dataset, scoring near-baseline accuracies with approximately $97\%$ of the pixels masked. This is rather surprising, as the images of this dataset are quite low-resolution, meaning that a classifier doesn't need a lot of pixels to perform adequately.   

In the first three datasets, the tradeoff between Fidelity and Interpretability appeared almost nonexistent. All models managed to \textit{hide} a large portion of their input, while not suffering with respect to classification performance. This will be discussed further in Sec. \ref{sec:discussion}.

CIFAR100 is tougher as a dataset, as it consists $100$ classes. This is evident by the diminished performance. Here the trade-off seems a bit tougher to overcome, as the models' Fidelity needed to drop by a bit $10\%$ to manage to \textit{hide} a significant portion of the input. 


\section{Discussion}
\label{sec:discussion}

\subsection{Relation to other approaches}

As mentioned in Section \ref{sec:related}, HnS is conceptually similar to ``occlusion sensitivity" analysis \cite{zeiler2014visualizing}. Through this technique the authors were able to visualize the features extracted by a fully trained CNN. However, this technique involves multiple inferences on a the model to identify the strongest feature map. While these techniques can theoretically achieve an arbitrarily high level of Interpretability, this would require performing a significant number of occlusions (as much as $2^M$, where $M$ are the input image's pixels). Needless to say this is computationally infeasible.

In contrast, HnS proposes using a trainable Neural Network (i.e. the hider) that learns what are the most relevant pixels for classification. This allows the hider to mask an ever-larger part of the input image, until the point where the least possible information is passed to the seeker. The downside to this is that the hider does require training; however, when trained it can make a prediction and generate a binary input mask in a single forward pass. This is arguably much more practical for real world applications. 

\subsection{Applications}
The HnS framework has several properties that can be exploited for various applications. These include:

\begin{itemize}
    \item \textit{Training fully-interpretable Neural Networks}. This has been discussed extensively throughout this paper.
    \item \textit{Training ``Student" networks}. Teacher-Student training \cite{hinton2015distilling}
    techniques have shown a lot of promise lately. A fully-trained hider could be used to train a much smaller CNN for a task which would normally require a model of a higher capacity. The intuition is that the hider has already learned to identify the important parts of an input image and by masking the irrelevant, it would help focus the CNN's attention to the parts that actually matter.
    \item \textit{Identifying bias in a dataset}. There have been examples of models ``cheating" in image classification by exploiting aspects of the images that humans would ignore (e.g. a watermark appearing in a corner is some class). By analyzing the saliency of a trained HnS, possible sources of bias in the data could be identified.
\end{itemize}


\subsection{Fidelity-Interpretability Tradeoff}
\label{sec:fir}

The tradeoff was not as costly as expected, most of the models that didn't collapse managed to hide a significant percentage.

\begin{figure}[h]
    \centering
    \includegraphics[width=0.6\linewidth]{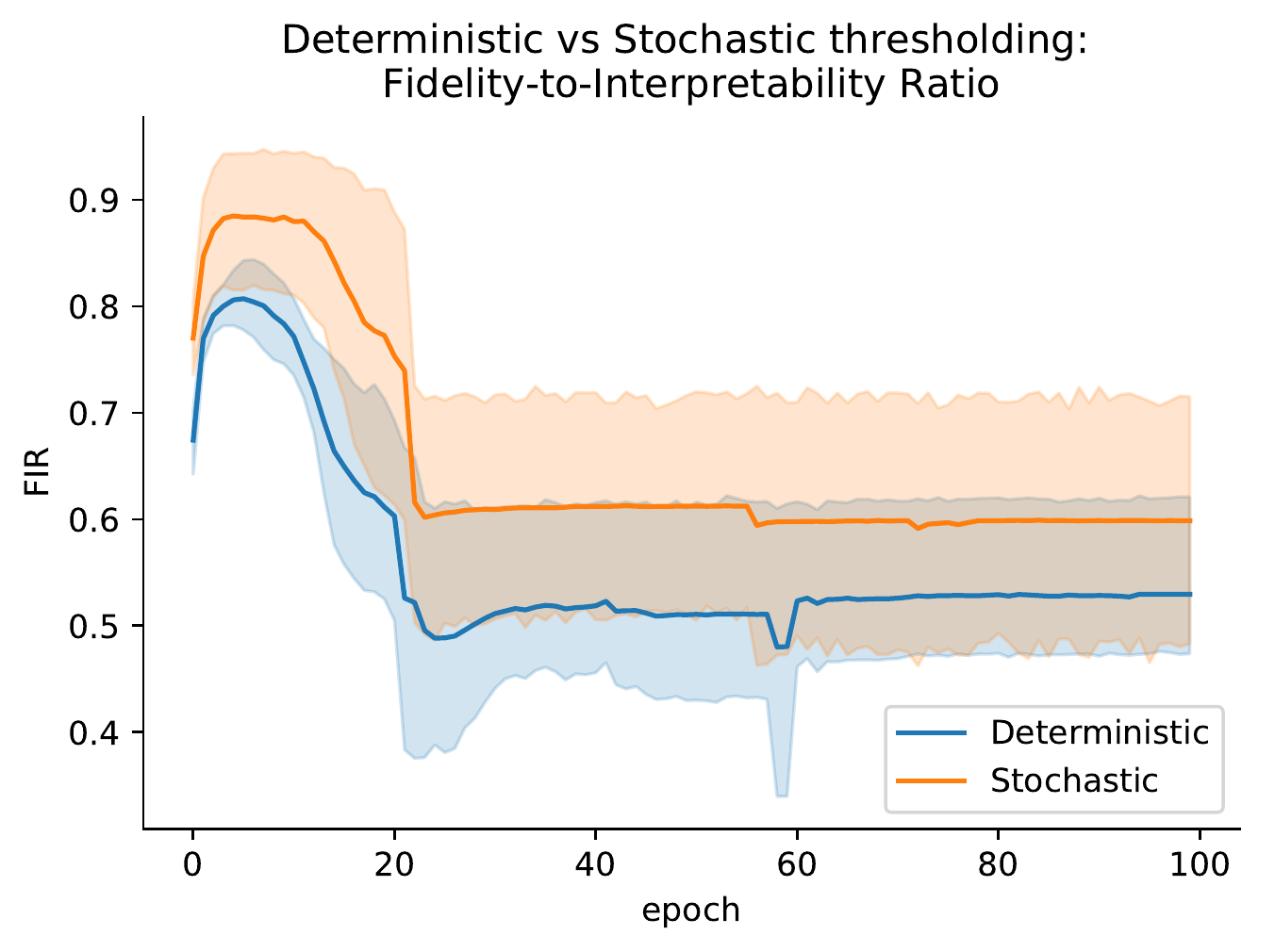}
    \caption{Fidelity-to-Interpretability (FIR) ratio of deterministic and stochastic models.}
    \label{fig:fir}
\end{figure}

Key insights can be discovered by examining the FIR for the HnS during training (Fig. \ref{fig:fir}). Both types of models, during the early steps tend to weigh in favor of Fidelity. This is natural due to the nature of the adaptive weighting (Sec. \ref{sec:adaptive}), which pushes the models to favour classification performance over masking. As the models start getting better at classifying and $\alpha$ starts dropping, the models start increasing their masking performance (i.e. their interpretability), which drops their FIR. During the latter stages of training, the models converge to their final FIR. 

\subsection{Deterministic vs Stochastic thresholding}

As analyzed in Section \ref{sec:det_vs_stoch}, deterministic and stochastic thresholding have very different effects to the training of the HnS. Stochastic models tend to settle for solutions with a higher degree of Fidelity. The stochasticity of the input mask leads to a better performing and more robust seeker. Additionally, it's very hard for a model to collapse to a solution where the hider masks the whole input image. This can be explained by the nature of the stochastic hider, which allows for a greater degree of exploration during training. Finally, these models exhibit a relatively low inter-model variance. This can, again, be attributed to the nature of stochastic thresholding. 

Deterministic models offer a higher degree of exploitation, especially during the generation of the input mask. This results in more ``extreme" solutions where the models either collapse or achieve optimally. This is evident by the fact that the ``best model" for $3$ out of $4$ datasets was a deterministic one. They are less robust than their stochastic counterparts, but manage to outperform them in terms of Interpretability, by sacrificing a bit of Fidelity. 

The differences between deterministic and stochastic models, with respect to their Fidelity and Interpretability can be seen in Figure \ref{fig:det_stoch_2d}. Deterministic models, seem to favor obtaining high Interpretability, while stochastic ones favor Fidelity.

The high degree of inter-model variance they exhibit compared to stochastic thresholding (Fig. \ref{fig:det_stoch_mask}), can be attributed to their nature. By utilizing a binary threshold, there could at any given point a set of weights ($W$ in Eq. \ref{eq:bdn}) that lead to activations near the threshold. This means that slight adjustments to those weights, during training, could result in large output fluctuations. Stochastic thresholding is superior in this regard as it ensures more stable transitions as the weights are updated. Small changes in the activations will only lead to small changes over the probability of a neuron being $0$ or $1$. Statistically, in the image, the same amount of pixels will be hidden or passed.

\section{Conclusion}

This paper proposes a new framework for increasing the interpretability of any Neural Network (NN), denoted the \textit{seeker}. It involves the use of another NN (called the \textit{hider}), which is trained to hide portions of its input. These two models are jointly trained to minimize classification error and maximize the percentage of the input that is hidden. As a result, the hider learns to recognize which parts of the input are possibly ``more interesting" and mask the rest.
This framework can be adapted for nearly any application, from Natural Language Processing (where the hider is a sequence-to-sequence model tasked at masking words), to Computer Vision (where the hider is an image-to-image model, e.g. an Autoencoder or a U-Net) and even structured data.

To achieve both goals of classifying accurately and masking a large portion of the input, the loss function is comprised of two components, which need to be regulated so that the model doesn't emphasize on only one of the goals. An adaptive weighting scheme of the two components is proposed as an alternative to manually tweaking their relative importance. 


The notions of Fidelity and Interpretability are introduced to help define and measure the two goals. Relevant literature describes the relationship of the above two as a trade-off. This claim was thoroughly investigated and is shown to be misleading, as we were able to achieve a high degree of Interpretability, while maintaining near-baseline classification performance. 

An extensive examination was conducted, regarding the best means of masking the input during training. Both deterministic and stochastic techniques for generating the mask were considered. While these two converged to roughly the same solutions, they achieved them with different means.

Experiments were performed on four different image classification datasets and proved that the HnS framework can be successfully applied to multiple tasks, without any fine-tuning.




\bibliographystyle{unsrt}
\bibliography{references.bib}

\end{document}